 \documentclass[final,3p,times]{elsarticle}

\usepackage{amsmath,amssymb}
\usepackage{graphicx}
\usepackage{booktabs}
\usepackage{xcolor}
\usepackage{tabularx}
\usepackage{multirow}
\usepackage{array}
\usepackage{subcaption}
\usepackage{float}
\usepackage{placeins}
\usepackage{needspace}
\usepackage{bm}

\usepackage[colorlinks=true,
            linkcolor=blue,
            citecolor=blue,
            urlcolor=blue]{hyperref}

\journal{Computer Methods in Applied Mechanics and Engineering}

\begin{document}

\begin{frontmatter}




\title{Architecture-Aware Mixed Precision for Replica-Exchange Training of Geometry-Aware Neural Operators with Uncertainty Quantification}


\author[label1]{Ujin Jeon}
\author[label2]{Michael 
Penwarden}
\author[label3]{Wenqian Chen}
\author[label4]{Guang Lin\corref{cor1}}
\ead{guanglin@purdue.edu}
\cortext[cor1]{Corresponding author}
\affiliation[label1]{organization={Department of Electrical and Computer 
            Engineering, Purdue University},
            city={West Lafayette},
            postcode={47907}, 
            state={IN},
            country={USA}}
\affiliation[label2]{organization={Sandia National Laboratories},
            addressline={1515 Eubank Blvd SE}, 
            city={Albuquerque},
            postcode={87123}, 
            state={NM},
            country={USA}}
\affiliation[label3]{organization={Pacific Northwest National Laboratory},
            addressline={3335 Innovation Blvd}, 
            city={Richland},
            postcode={99354}, 
            state={WA},
            country={USA}}
\affiliation[label4]{organization={Department of Mathematics and Mechanical 
            Engineering, Purdue University},
            city={West Lafayette},
            postcode={47907}, 
            state={IN},
            country={USA}}

\begin{abstract}
Geometry-aware neural operators face substantial computational and memory demands when learning across complex physical domains. We introduce Architecture-Aware Mixed Precision (AA-MP), a replica-exchange training strategy that allocates numerical precision according to replica roles and architecture-specific operations. By retaining FP32 for refinement while selectively employing BF16 for exploration, AA-MP preserves predictive accuracy while improving training efficiency. Across fluid dynamics and three-dimensional structural mechanics benchmarks, AA-MP achieves comparable accuracy to full-precision training over an ensemble of runs, while reducing hot-replica peak GPU memory by 46.9\% and training time by up to 25.8\%. It also outperforms naive Automatic Mixed Precision (AMP) in memory efficiency and computational cost, with increasing memory advantages at larger model and geometry scales. Ablation studies show that retaining FP32 in the cold-replica is important for predictive accuracy, while additional FP32 operations in the exploratory replica provide limited benefits relative to their computational cost. Finally, conformal prediction for uncertainty calibration shows that AA-MP achieves near-nominal coverage with narrower prediction intervals than the alternatives. These results advance the scalability of geometry-aware scientific machine learning by reducing computational costs without compromising predictive accuracy or uncertainty quantification.

\end{abstract}




\end{frontmatter}




\section{Introduction}
\label{sec:introduction}

Scientific machine learning (SciML) has become an important tool for accelerating simulation, inference, and prediction in computational physics. Methods such as physics-informed neural networks (PINNs), Deep Operator Networks (DeepONets), Fourier neural operators (FNOs), and related neural operator architectures have shown strong potential for learning parametric solution operators and constructing fast surrogate models for systems governed by partial differential equations (PDEs) \cite{raissi2019physics,karniadakis2021physics,lu2021learning,li2021fourier,kovachki2023neural}. These methods are especially useful in many-query settings, including design optimization, uncertainty quantification, inverse modeling, and real-time prediction, where repeated high-fidelity PDE solves are computationally expensive.

A central challenge is extending neural operators to complex and variable geometries. DeepONet supports evaluation at arbitrary spatial query locations, while conventional FNO formulations commonly use structured discretizations. Nevertheless, representing complex and varying geometries remains challenging and may require additional geometric encoding or architecture-specific treatment \cite{shukla2024deep}. The Arbitrary Geometry-encoded Transformer (ArGEnT) addresses this challenge by representing geometry through point-cloud coordinates and signed-distance information and using attention to construct geometry-conditioned solution features \cite{chen2026argentarbitrarygeometryencodedtransformer}. In this work, we use the cross-attention ArGEnT formulation, in which sampled query locations form the query representation while a separate geometry point cloud provides the key and value representations. ArGEnT is integrated as the geometry-aware trunk of a DeepONet-style architecture, while the branch network encodes the non-geometric physical and operating parameters.

This geometric flexibility introduces substantial computational cost. Geometry-aware Transformer operators repeatedly process sampled query locations and geometry point clouds through projection, attention, and feedforward operations, making GPU memory consumption and training throughput important practical constraints. Mixed-precision computation provides a natural mechanism for reducing these costs and has been studied extensively in deep learning, numerical linear algebra, scientific computing, and, more recently, neural operator learning \cite{micikevicius2018mixed,abdelfattah2021survey,tu2024guaranteed}.

Existing mixed-precision strategies often assign numerical precision according to operation type, numerical sensitivity, model component, or backend execution policy. In this work, we additionally use the algorithmic role of each replica in replica-exchange optimization to determine its precision policy. The low-temperature replica is responsible for refinement and final model evaluation, whereas the high-temperature replica is used to promote broader exploration of the optimization landscape. This asymmetry motivates treating numerical precision as a replica-dependent computational resource rather than applying the same precision policy to both chains.

Accordingly, we propose an architecture-aware mixed-precision (AA-MP) strategy for replica-exchange training of geometry-aware neural operators. The low-temperature cold replica uses 32-bit floating-point (FP32) computation for refinement and model evaluation, whereas the high-temperature exploratory replica uses a manually controlled bfloat16 (BF16) model representation for its forward and backward computation. The hot-replica preconditioned reSGLD update retains FP32 master parameters and first- and second-moment states, while the DeepONet latent reduction and scalar prediction-error and exchange-energy reductions are also evaluated in FP32. Replica-exchange proposals couple this reduced-precision exploratory replica to the full-precision refinement replica. AA-MP therefore incorporates mixed precision directly into the exploration--refinement structure of replica exchange rather than applying a uniform precision policy to both replicas. Throughout this work, the notation A/B denotes the precision policies of the cold and hot replicas, respectively.

The main contributions of this work are as follows.

\begin{itemize}

    \item \textbf{Replica-dependent mixed-precision training.}
    We introduce a mixed-precision replica-exchange strategy that assigns numerical precision according to replica role. The low-temperature cold replica is maintained entirely in FP32 for refinement and evaluation, whereas the high-temperature exploratory replica executes the ArGEnT--DeepONet forward and backward computation using the proposed architecture-aware mixed-precision mechanism.

    \item \textbf{Explicit precision control for preconditioned reSGLD.}
    We couple the BF16 exploratory model representation with FP32 master parameters and first- and second-moment optimizer states. The branch--trunk product is formed in BF16 and promoted to FP32 before the latent reduction, while the prediction error and exchange-energy reductions are evaluated in FP32. This explicit precision map separates reduced-precision neural-network computation from FP32 optimizer and reduction operations without relying on automatic mixed-precision (AMP) autocasting.

    \item \textbf{Controlled precision-allocation and protection ablations.}  
    We evaluate the proposed FP32/AA-MP configuration over 10 independent seeds on the AirfRANS fluid-dynamics and jet-engine bracket structural-mechanics benchmarks, comparing it with FP32/FP32, FP32/AMP, and FP32/Naive-BF16. We further examine alternative cold/hot replica allocations and progressively increased FP32 protection within the exploratory replica to determine how precision placement affects predictive accuracy and computational efficiency.
 
    \item \textbf{Computational efficiency and scaling.}
    FP32/AA-MP reduces hot-phase peak allocated GPU memory by 46.9\% on both benchmarks and hot-replica step time by 25.0\% on AirfRANS and 25.8\% on the bracket benchmark relative to FP32/FP32, while maintaining comparable predictive accuracy. Relative to FP32/AMP, the proposed configuration further reduces GPU memory by approximately 9.4\% and hot-replica step time by 5.5\% and 3.2\% on AirfRANS and the bracket benchmark, respectively. Scaling experiments further show that AA-MP maintains lower GPU-memory requirements than standard AMP and the available FP32 reference as geometry point-cloud size and Transformer embedding dimension increase, with larger absolute savings at the highest tested workloads.
    
    \item \textbf{Uncertainty quantification under different precision strategies.}
    Using predictions from retained cold-replica states, we evaluate post-training uncertainty quantification on AirfRANS for FP32/FP32, FP32/AMP, FP32/AA-MP, and FP32/Naive-BF16. We apply conformal risk control and compare empirical case-averaged spatial coverage, prediction-interval width, and interval scores to assess uncertainty calibration and interval sharpness under the different precision strategies.

\end{itemize}

The remainder of the paper is organized as follows. Section~\ref{sec:background} reviews mixed-precision computation, ArGEnT--DeepONet, and replica-exchange SGLD. Section~\ref{sec:methodology} presents the proposed AA-MP replica-exchange framework and its precision-aware preconditioned reSGLD implementation. Section~\ref{sec:experiments} presents the benchmark results, precision-allocation ablations, computational-efficiency studies, and post-training conformal-UQ evaluation. Section~\ref{sec:conclusion} summarizes the main findings and discusses future directions.

\section{Related Work and Background}
\label{sec:background}
This section reviews prior work on mixed-precision computation relevant to scientific machine learning and introduces the two technical components underlying the proposed methodology: ArGEnT for geometry-aware operator learning and replica-exchange SGLD for temperature-based exploration. Together, these components provide the architectural and optimization framework on which the proposed replica-dependent precision strategy is built.

\subsection{Mixed-Precision Training for Scientific Machine Learning}

Mixed-precision computation has been widely used to reduce the memory and computational cost of machine-learning models. In deep learning, mixed-precision training was popularized through reduced-precision forward and backward operations, high-precision master weights, and loss scaling to
avoid gradient underflow \cite{micikevicius2018mixed}. Related approaches have been developed for neural ordinary differential equations and mixed-precision quantization, where precision is assigned according to stability, sensitivity, entropy, or accuracy-aware criteria
\cite{celledoni2026mixed,bablani2024efficient,sun2022entropy,qin2025mixed}.

Mixed precision also has a long history in numerical linear algebra and high-performance scientific computing. Classical mixed-precision iterative refinement performs expensive algebraic operations in reduced precision while using higher precision for residual correction, enabling low-precision hardware acceleration while recovering higher-accuracy solutions \cite{langou2006exploiting,kurzak2007implementation,higham2022mixed,
abdelfattah2021survey}. Modern GPU implementations exploit tensor cores and other reduced-precision hardware to accelerate dense and sparse linear algebra, Krylov methods, eigensolvers, geostatistical modeling, weather prediction, and large-scale scientific simulations
\cite{haidar2018harnessing,haidar2020mixed,hogg2010fast, idomura2020acceleration,abdulah2022accelerating,maynard2019mixed,
yamazaki2015mixed}. These studies demonstrate that numerical precision can serve as an algorithmic design parameter for balancing computational efficiency and numerical accuracy.

Mixed precision has also been investigated directly in scientific machine learning. Hayford et al.~\cite{hayford2024mixedprecision} studied mixed-precision training for physics-informed neural networks (PINNs) and DeepONets, demonstrating reductions in computational cost and memory usage while examining numerical difficulties associated with low-precision computation. Related work on mixed-precision neural operators has also investigated approximation guarantees and numerical precision effects~\cite{tu2024guaranteed}.

These studies motivate the use of mixed precision in scientific machine learning, while leaving room to investigate how precision allocation can be tailored to specific neural operator architectures and optimization procedures. In this work, we investigate architecture-aware and replica-dependent precision allocation for ArGEnT--DeepONet training with preconditioned replica-exchange SGLD. The low-temperature refinement replica is retained in FP32, whereas the high-temperature exploratory replica uses an explicitly controlled BF16 model representation with FP32 master parameters, optimizer states, and selected numerical reductions. By directly comparing this configuration with standard BF16 AMP under the same replica-exchange training framework, we evaluate whether explicit precision control provides additional computational savings while maintaining competitive predictive accuracy.

\subsection{Arbitrary Geometry-encoded Transformer for Operator Learning (ArGEnT)}

The Arbitrary Geometry-encoded Transformer (ArGEnT) architecture was proposed in \cite{chen2026argentarbitrarygeometryencodedtransformer}. ArGEnT is a geometry-aware neural operator architecture designed to learn solution operators on complex and varying physical domains. DeepONet supports evaluation at arbitrary spatial query locations, while conventional Fourier neural operator (FNO) formulations commonly use structured discretizations. Representing complex and varying geometries nevertheless requires suitable geometric inputs or architectural mechanisms. ArGEnT addresses this challenge by incorporating geometry through point-based representations and decoupling geometry encoding from query-point evaluation. In this work, ArGEnT serves as the baseline geometry-aware operator-learning model to which mixed-precision training is applied.

Let $x \in \mathbb{R}^{d}$ denote a spatial query point and let $\phi(x)$ denote the signed distance function (SDF) value associated with the geometry. We write the augmented query input as
\begin{equation}
    \tilde{x} = (x,\phi(x)),
\end{equation}
where additional pointwise features can be appended when available. The geometry is represented by a point cloud
\begin{equation}
    \Omega = \left\{ \tilde{x}^{g}_{i} \right\}_{i=1}^{M},
    \qquad
    \tilde{x}^{g}_{i} = \left(x^{g}_{i},\phi(x^{g}_{i})\right),
\end{equation}
where $x^{g}_{i}$ is a geometry point and $\phi(x_i^g)$ is the corresponding SDF value. This representation allows the model to map complex domains without explicit finite-dimensional geometric parameterization.

In this work, we use the cross-attention ArGEnT formulation to evaluate the proposed mixed-precision training strategy.

In the cross-attention variant, the query points and geometry representation are decoupled. The embedded query features form the query matrix $Q$, whereas the embedded geometry point cloud $\Omega$ provides the key and value representations $K$ and $V$. For a single attention head, let $Q\in\mathbb{R}^{N_q\times d}$ and $K,V\in\mathbb{R}^{N_{\mathrm{attn}}\times d}$, where $N_q$ is the number of query points, $N_{\mathrm{attn}}$ is the number of geometry tokens, and $d$ is the head dimension. The Galerkin-type cross-attention operation is expressed as
\begin{equation}
    \operatorname{Attn}(Q,K,V)
    =
    \frac{1}{N_{\mathrm{attn}}}
    Q\left(\widetilde{K}^{\top}\widetilde{V}\right),
    \label{eq:argent_kernel_attention}
\end{equation}
where $\widetilde{K}$ and $\widetilde{V}$ denote the normalized key and value representations, respectively. The operation is softmax-free: the key--value aggregation is performed before multiplication by the query representation, without applying softmax normalization to $QK^{\top}$. The resulting cross-attention operator can be written abstractly as
\begin{equation}
    \mathcal{G}_{\mathrm{CA}}:
    \left(\tilde{x},\Omega\right)
    \mapsto
    f(\tilde{x},\Omega).
\end{equation}
Because the query points and geometry representation are supplied separately, the solution can be evaluated at query locations sampled independently of the geometry point cloud.

For physical systems whose solution depends both on geometry and on non-geometric parameters, ArGEnT is integrated into a DeepONet-style branch--trunk architecture. Let $\mu$ denote the non-geometric physical parameters, such as operating conditions or applied loads. For a problem with $d_{\mathrm{out}}$ predicted physical fields and latent dimension $J$, the branch network produces field-dependent coefficients
\begin{equation}
    c(\mu)
    =
    \left\{
    c_{\ell j}(\mu)
    \right\}_{\ell=1,\ldots,d_{\mathrm{out}},\,j=1,\ldots,J},
\end{equation}
while the ArGEnT trunk produces geometry-conditioned latent features
\begin{equation}
    f(\tilde{x},\Omega)
    =
    \left\{
    f_{\ell j}(\tilde{x},\Omega)
    \right\}_{\ell=1,\ldots,d_{\mathrm{out}},\,j=1,\ldots,J}.
\end{equation}
The predicted value of output field $\ell$ is obtained by an elementwise branch--trunk product followed by reduction over the latent dimension,
\begin{equation}
    \hat{u}_{\ell}(\tilde{x};\Omega,\mu)
    =
    \sum_{j=1}^{J}
    c_{\ell j}(\mu)
    f_{\ell j}(\tilde{x},\Omega),
    \qquad
    \ell=1,\ldots,d_{\mathrm{out}}.
    \label{eq:argent_deeponet_output}
\end{equation}
The vector-valued prediction is therefore
$\hat{\mathbf{u}}
=
(\hat{u}_1,\ldots,\hat{u}_{d_{\mathrm{out}}})$.

The model is trained in a supervised operator-learning setting using high-fidelity simulation data. Given vector-valued reference solutions $\mathbf{u}(x_i;\Omega,\mu)$ at query points $\{x_i\}_{i=1}^{N}$, the baseline supervised objective is the mean-squared error (MSE),
\begin{equation}
    \mathcal{L}_{\mathrm{MSE}}
    =
    \frac{1}{N}
    \sum_{i=1}^{N}
    \left\|
    \hat{\mathbf{u}}(\tilde{x}_i;\Omega,\mu)
    -
    \mathbf{u}(x_i;\Omega,\mu)
    \right\|_{2}^{2}.
    \label{eq:supervised_mse}
\end{equation}

\FloatBarrier

\subsection{Preconditioned Replica Exchange SGLD}

Stochastic gradient Langevin dynamics (SGLD) was introduced as a scalable stochastic-gradient sampling method for Bayesian learning \cite{welling2011bayesian}. Let $\theta_k$ denote the trainable parameters at iteration $k$ and let $U_{\mathcal{B}_k}(\theta_k)$ denote the stochastic training energy evaluated on minibatch $\mathcal{B}_k$. To motivate the optimization dynamics used in this work, consider the diagonally preconditioned update~\cite{li2016preconditionedsgld}
\begin{equation}
\label{eq:psgld_general}
    \theta_{k+1}
    =
    \theta_k
    -
    \eta_k G_k
    \nabla U_{\mathcal{B}_k}(\theta_k)
    +
    \sqrt{2\eta_k T}\,
    G_k^{1/2}\xi_k,
    \qquad
    \xi_k \sim \mathcal{N}(0,I),
\end{equation}
where $\eta_k$ is the step size, $T$ is the Langevin temperature, and $G_k$ is a positive diagonal preconditioner that rescales both the stochastic-gradient update and Langevin perturbation. Equation~\eqref{eq:psgld_general} provides a reference preconditioned Langevin update. The moment-based adaptive preconditioner and stochastic optimization updates used in this work are specified in Section~\ref{sec:methodology}; the reference expression is not taken to establish exact invariant-distribution sampling for the adaptive implementation.

Replica exchange, also known as parallel tempering or exchange Monte Carlo, improves exploration by evolving multiple chains at different temperatures and periodically proposing swaps between them \cite{hukushima1996exchange}. For two chains with temperatures $T_c<T_h$ and states $\theta^c$ and $\theta^h$, the exchange is accepted with probability
\begin{equation}
\alpha
=
\min\left\{
1,
\exp\left[
\left(
\frac{1}{T_c}
-
\frac{1}{T_h}
\right)
\left(
U(\theta^c)-U(\theta^h)
\right)
\right]
\right\}.
\label{eq:ideal_replica_exchange}
\end{equation}
Equation~\eqref{eq:ideal_replica_exchange} gives the ideal replica-exchange acceptance probability based on the full training energy. The high-temperature replica promotes broader exploration of the loss landscape, whereas the low-temperature replica provides lower-temperature refinement. Replica-exchange stochastic-gradient methods extend this principle to stochastic-gradient learning problems, where non-convexity and minibatch noise complicate the optimization landscape~\cite{deng2020nonconvex,li2026fastresgld}. In this work, replica exchange additionally provides the algorithmic structure on which replica-dependent numerical precision is assigned; the stochastic minibatch exchange criterion used in the actual implementation is introduced in Section~\ref{sec:methodology}. 

\section{Methodology}
\label{sec:methodology}

\begin{figure}[H]
    \centering
    \includegraphics[width=\textwidth]
    {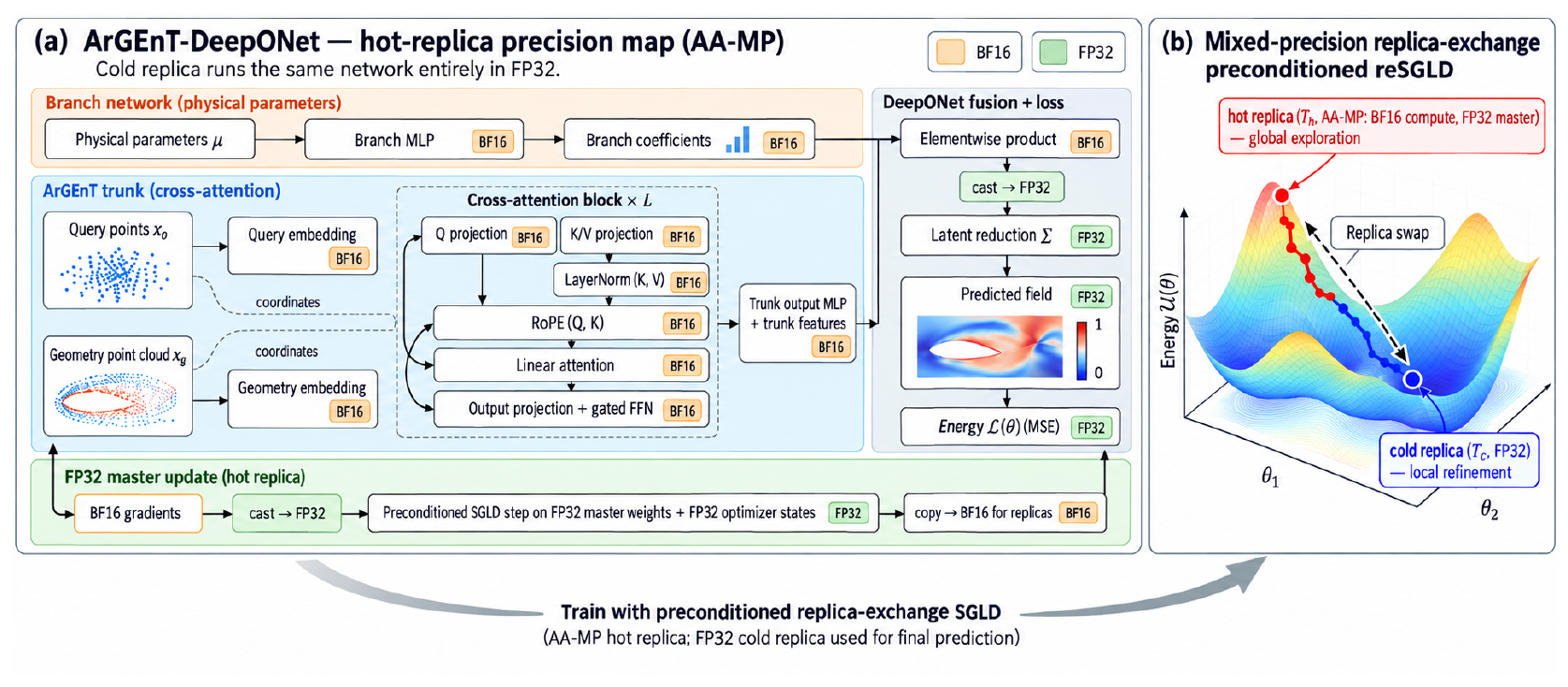}
    \caption{Overview of the proposed AA-MP ArGEnT--DeepONet training framework. (a) The DeepONet branch encodes non-geometric physical parameters, while the cross-attention ArGEnT trunk uses query features as queries and the geometry point cloud as keys and values. The exploratory hot replica uses a manually controlled BF16 live model, while the latent reduction, prediction-error and energy reductions, master parameters, and optimizer states are maintained in FP32. The cold replica executes the same network entirely in FP32. (b) Two replicas are evolved using preconditioned reSGLD at $T_c<T_h$. The higher-temperature hot replica promotes broader exploration, whereas the FP32 cold replica performs low-temperature refinement and model evaluation. Replica-exchange proposals periodically exchange the parameter and associated optimizer states between the two replica slots.}
    \label{fig:aamp_framework}
\end{figure}

We present a replica-dependent mixed-precision strategy for training ArGEnT--DeepONet models with preconditioned replica-exchange stochastic gradient Langevin dynamics (reSGLD). Rather than applying the same automatic mixed-precision policy to both replicas, the proposed method assigns numerical precision according to replica role. The low-temperature cold replica remains entirely in FP32 for refinement and model evaluation, whereas the high-temperature exploratory replica uses a manually controlled BF16 live model for its computationally intensive forward and backward operations. The hot-replica optimizer retains FP32 master parameters and first- and second-moment states, while the DeepONet latent reduction and scalar prediction-error and exchange-energy reductions are evaluated in FP32. This design couples numerical precision directly to the exploration--refinement structure of replica exchange.

\subsection{Mixed Precision Replica Exchange}

The proposed method incorporates mixed precision into the replica-exchange training strategy rather than treating reduced precision as a uniform acceleration mechanism. The cold chain is evaluated entirely in FP32 and acts as the low-temperature refinement and evaluation replica. The hot chain operates at a higher Langevin temperature and uses reduced-precision computation for the expensive ArGEnT--DeepONet forward and backward passes. Exchange proposals allow parameter states discovered by the exploratory replica to enter the cold replica while the cold slot subsequently continues its updates in FP32. The resulting strategy therefore assigns reduced precision to the exploratory replica while retaining full precision in the refinement replica.

In the present optimization setting, $\widehat{U}_{\mathcal{B}}(\theta)$ denotes the minibatch supervised training objective used by the implementation. The temperatures $T_c$ and $T_h$ are algorithmic hyperparameters that control the Langevin perturbation magnitude and the exchange rule.

Let $\theta$ denote the trainable parameters of the ArGEnT-DeepONet operator. We evolve two replicas using preconditioned replica-exchange stochastic gradient Langevin dynamics (reSGLD) at Langevin temperatures $T_c<T_h$, where $T_c$ and $T_h$ denote the cold- and hot-replica temperatures, respectively. The cold chain is updated using full-precision evaluations. Its stochastic gradient is therefore written as
\begin{equation}
g_k^c
=
\nabla \widehat{U}_{32,\mathcal{B}_k}(\theta_k^c),
\end{equation}
where $\widehat{U}_{32,\mathcal{B}_k}$ denotes the training objective on minibatch $\mathcal{B}_k$ evaluated in full precision. For the hot chain, let $\theta_k^h$ denote the FP32 master parameters and let
\begin{equation}
\theta_{k,\mathrm{live}}^h
=
\operatorname{cast}_{\mathrm{BF16}}(\theta_k^h)
\end{equation}
denote the synchronized BF16 live-model parameters. The hot-replica gradient used by the optimizer is
\begin{equation}
g_k^h
=
\operatorname{cast}_{32}
\left[
\operatorname{grad}_{\mathrm{live}}
\widehat{U}_{\mathrm{mp},\mathcal{B}_k}
(\theta_{k,\mathrm{live}}^h)
\right],
\end{equation}
where $\operatorname{grad}_{\mathrm{live}}$ denotes the gradient computed by automatic differentiation with respect to the BF16 live-model parameters. The resulting gradient is promoted to FP32 before updating the master parameters. This notation describes the implemented mixed-precision gradient computation rather than differentiation through the FP32-to-BF16 casting operation.

For each replica, the preconditioned reSGLD optimizer maintains FP32 first- and second-moment buffers. Let $r\in\{c,h\}$ denote the cold or hot replica, and let $g_k^r$ denote its stochastic gradient at update $k$. We index the optimizer updates by $k\geq 1$ and initialize the moment buffers as $m_0^r=v_0^r=0$. The first- and second-moment estimates are updated as
\begin{align}
m_k^r
&=
\beta_1 m_{k-1}^r
+
(1-\beta_1)g_k^r,
\label{eq:first_moment}
\\
v_k^r
&=
\beta_2 v_{k-1}^r
+
(1-\beta_2)
\left(g_k^r \odot g_k^r\right),
\label{eq:second_moment}
\end{align}
where $\odot$ denotes elementwise multiplication. The corresponding bias-corrected moment estimates are
\begin{equation}
\widehat{m}_k^r
=
\frac{m_k^r}{1-\beta_1^k},
\qquad
\widehat{v}_k^r
=
\frac{v_k^r}{1-\beta_2^k}.
\label{eq:bias_corrected_moments}
\end{equation}

Using the second-moment estimate, we define the adaptive diagonal preconditioner as
\begin{equation}
G_k^r
=
\operatorname{diag}
\left[
\min\left(
G_{\max},
\frac{1}{\sqrt{\widehat{v}_k^r}+\epsilon}
\right)
\right],
\qquad
r\in\{c,h\},
\label{eq:adaptive_preconditioner}
\end{equation}
where the minimum is applied elementwise. The constant $G_{\max}$ bounds the magnitude of the diagonal preconditioning coefficients and prevents excessively large updates in coordinates with very small estimated second moments.

Using the adaptive diagonal preconditioner in
Eq.~\eqref{eq:adaptive_preconditioner}, we define the cold- and hot-replica updates as
\begin{align}
\theta_{k+1}^{c}
&=
\theta_k^{c}
-
\eta_k G_k^{c}\widehat{m}_k^{c}
+
\sqrt{2\eta_k T_c}
\left(G_k^{c}\right)^{1/2}
\xi_k^{c},
\qquad
\xi_k^{c}\sim\mathcal{N}(0,I),
\label{eq:cold_resgld_update}
\\
\theta_{k+1}^{h}
&=
\theta_k^{h}
-
\eta_k G_k^{h}\widehat{m}_k^{h}
+
\sqrt{2\eta_k T_h}
\left(G_k^{h}\right)^{1/2}
\xi_k^{h},
\qquad
\xi_k^{h}\sim\mathcal{N}(0,I).
\label{eq:hot_resgld_update}
\end{align}
Here the square root of the diagonal preconditioner acts elementwise. For the cold chain, $g_k^c$ is obtained from an FP32 forward and backward pass. In AA-MP, the live hot-replica model parameters are stored in BF16 and the hot forward and backward passes are executed directly in the manually assigned BF16 representation without AMP autocasting. The resulting hot-replica gradients are promoted to FP32 for the preconditioned reSGLD update, whose master parameters and first- and second-moment states are maintained in FP32. After each optimizer step, the updated FP32 master parameters are cast and synchronized back to the BF16 live model. The precision used for the branch--trunk reduction and exchange-energy evaluation is described in Section~\ref{sec:aamp_precision_policy} and summarized in Table~\ref{tab:precision_policy}.

Every $K_{\mathrm{swap}}$ iterations after the burn-in period, we propose an exchange between the cold- and hot-replica states. The same Langevin temperatures $T_c$ and $T_h$ used in the preconditioned reSGLD dynamics are used in the exchange criterion.

Let $\widehat{U}_{\mathcal{B}}^{c}$ and $\widehat{U}_{\mathcal{B}}^{h}$ denote the minibatch estimates of the
training energy obtained from the current cold- and hot-replica model representations, respectively. We define the stochastic exchange criterion as
\begin{equation}
\widehat{\alpha}
=
\min\left\{
1,
\exp\left[
\left(
\frac{1}{T_c}
-
\frac{1}{T_h}
\right)
\left(
\widehat{U}_{\mathcal{B}}^{c}(\theta^c)
-
\widehat{U}_{\mathcal{B}}^{h}(\theta^h)
\right)
\right]
\right\}.
\label{eq:stochastic_exchange}
\end{equation}
For each exchange proposal, the cold- and hot-replica energies are recomputed on the same current minibatch using the active numerical representation of each replica. The cold replica performs an FP32 forward evaluation, whereas the AA-MP hot replica performs its forward evaluation using the manually controlled BF16 model representation without AMP autocasting. For AA-MP, the branch--trunk product is formed in BF16 and subsequently promoted to FP32 before the latent reduction. The resulting prediction and the squared-error reduction used to form the minibatch energy are therefore evaluated in FP32. The scalar exchange criterion in
Eq.~\eqref{eq:stochastic_exchange} is then evaluated in FP32 from the current cold- and hot-replica minibatch energies. Thus, the computationally intensive hot-replica model evaluation remains in BF16 while the final
DeepONet reduction and scalar exchange calculation are performed in FP32.

An exchange is accepted with probability $\widehat{\alpha}$. Upon acceptance, the parameter state and the associated preconditioned reSGLD optimizer state are exchanged between the two replicas. Maintaining the optimizer state together with its parameter vector preserves the correspondence between the parameters and the adaptive preconditioner. After an accepted exchange, the cold replica continues its subsequent
updates in FP32, while the hot-replica BF16 model representation is synchronized from its corresponding FP32 master parameters.

Equation~\eqref{eq:stochastic_exchange} defines the stochastic exchange rule used in our implementation. Unlike the exact Metropolis criterion for replica exchange, this rule uses minibatch energy estimates obtained from replicas evaluated at different numerical precisions. Consequently, the acceptance probability is approximate and does not guarantee detailed balance with respect to the full-data Gibbs distributions. We therefore interpret replica exchange as a mechanism for coupling exploratory and refinement trajectories, rather than as an exact sampling procedure.

This formulation differs from enabling standard AMP for the hot replica. In the AMP baseline, model parameters remain stored in FP32 while selected operations are executed in BF16 according to PyTorch autocast rules. In contrast, AA-MP explicitly represents the live hot-replica model in BF16 and executes its forward and backward computations without autocast, while retaining FP32 master parameters and first- and second-moment states for the preconditioned reSGLD update. The cold replica remains entirely in FP32. AA-MP therefore differs from AMP both in the representation of the exploratory replica and in the explicit separation between reduced-precision model computation and FP32 optimizer state.

\subsection{Architecture-Aware Mixed-Precision Policy}
\label{sec:aamp_precision_policy}
The proposed AA-MP strategy explicitly assigns numerical precision across the ArGEnT--DeepONet training pipeline rather than relying on automatic mixed-precision casting. The cold replica remains in FP32. In the exploratory hot replica, the live model parameters and major intermediate tensors use BF16, including those associated with the branch and trunk networks, embedding layers, Q/K/V projections, K/V normalization, rotary positional encoding (RoPE), kernel-attention operations, and feedforward layers. This precision policy specifies the tensor representations and explicitly controlled operations; backend kernels may perform internal accumulation at higher precision.

The DeepONet branch--trunk elementwise product is also formed in BF16 and is then promoted to FP32 before reduction over the latent dimension. The resulting prediction and prediction-error reduction are evaluated in FP32. During optimization, gradients produced by the BF16 live model are promoted to FP32, while the preconditioned reSGLD optimizer maintains FP32 master parameters and first- and second-moment states. The scalar energy reduction used by the replica-exchange procedure is likewise evaluated in FP32. Table~\ref{tab:precision_policy} summarizes this precision structure.

\begin{table}[htbp!]
\centering
\caption{Precision structure of the proposed AA-MP configuration.
The cold replica remains entirely in FP32. The table summarizes the
manually controlled precision policy of the exploratory hot replica.}
\label{tab:precision_policy}
\begin{tabular}{p{0.37\linewidth} p{0.17\linewidth} p{0.36\linewidth}}
\toprule
\textbf{Operation} &
\textbf{Precision} &
\textbf{Implementation role} \\
\midrule

Hot-replica live model parameters
&
BF16
&
Explicit reduced-precision model representation without autocast.
\\
\midrule
Branch/trunk embeddings and MLPs
&
BF16
&
Reduced-precision feature computation.
\\
\midrule
Q/K/V projections, K/V LayerNorm, and RoPE
&
BF16
&
Cross-attention and positional feature computation.
\\
\midrule
Kernel attention, output projection, and feedforward layers
&
BF16
&
Computationally intensive Transformer operations.
\\
\midrule
Branch--trunk elementwise product
&
BF16
&
DeepONet feature fusion before latent reduction.
\\
\midrule
Latent reduction and prediction
&
FP32
&
The BF16 branch--trunk product is promoted before summation.
\\
\midrule
Hot-replica gradients
&
BF16 $\rightarrow$ FP32
&
Gradients are promoted before the optimizer update.
\\
\midrule
Master parameters and optimizer moments
&
FP32
&
Preconditioned reSGLD parameter and moment states.
\\
\midrule
Prediction-error and energy reductions
&
FP32
&
Final scalar loss and exchange-energy reductions.
\\

\bottomrule
\end{tabular}
\end{table}

\FloatBarrier

\section{Numerical Experiments}
\label{sec:experiments}

\subsection{Experimental Setup}
\label{sec:experimental_setup}
All primary replica-exchange experiments were conducted on a single NVIDIA H100 80 GB GPU using PyTorch 2.5.1.post303 with a CUDA 12.0 build. The models were trained for 40,000 iterations with a case batch size of 32, a query-point batch size of 3,000, a geometry point-cloud size of 3,000, and an ArGEnT embedding dimension of 128. The initial learning rate was $3\times10^{-4}$.

The experiments use the implemented preconditioned reSGLD optimizer with cold- and hot-replica Langevin temperatures $T_c=10^{-6}$ and $T_h=10^{-5}$, respectively. The optimizer uses $\beta_1=0.9$ for the first-moment estimate, $\beta_2=0.99$ for the second-moment estimate, $\epsilon=10^{-8}$, and a maximum diagonal preconditioner value of $G_{\max}=50$. The learning rate is multiplied by $0.99$ every 100 completed iterations. Replica-exchange proposals are evaluated every 100 completed iterations after a 100-iteration burn-in. The same Langevin temperatures $T_c$ and $T_h$ used in the replica updates are used in the exchange criterion.

The primary resource comparison considers three precision strategies: FP32/FP32, FP32/AMP, and FP32/AA-MP. FP32/FP32 uses full-precision computation for both replicas. In the proposed FP32/AA-MP configuration, the cold replica remains in FP32 while the live hot-replica model is represented and evaluated in BF16 without autocast; the hot preconditioned reSGLD master parameters and moment states remain in FP32, and selected output and energy reductions are performed in FP32. In the FP32/AMP baseline, the hot-replica parameters remain stored in FP32 while BF16 execution is selected through PyTorch autocast. FP32/Naive-BF16 is additionally included as a reduced-precision ablation in the predictive-accuracy, scaling, and post-training UQ evaluations. For the primary predictive-accuracy and computational-efficiency comparisons, each precision configuration is evaluated over 10 independent seeds. The experiments are designed to compare precision policies within the same replica-exchange training framework rather than to establish an optimization advantage of replica exchange over unrelated single-chain optimizers.

For the jet-engine bracket experiments, 5,052 cases are used for training and 1,263 cases for testing. For AirfRANS, the experiments use the dataset with 800 training cases and 200 test cases. The primary predictive-accuracy and computational-efficiency comparisons use the same 40,000-iteration replica-exchange protocol for both benchmarks. The architecture-level precision-protection ablation uses a separate 20,000-iteration screening protocol, while the post-training conformal-UQ study uses models trained for 100,000 iterations followed by an additional Langevin sampling phase. These distinct experimental protocols are described in their respective subsections.

At each test evaluation, 3,000 query points are sampled without replacement from each test case. The sampled point set is regenerated at each evaluation rather than held fixed across training iterations. Predictive accuracy is quantified using the per-case
relative $\ell_2$ error. For test case $j$ and
predicted field $k$, we define
\begin{equation}
e_{j,k} =
\frac{
\left(
\sum_{i=1}^{N_q}
\left|
\widehat{u}_{j,k}(\mathbf{x}_i)
-
u_{j,k}(\mathbf{x}_i)
\right|^2
\right)^{1/2}
}{
\left(
\sum_{i=1}^{N_q}
\left|
u_{j,k}(\mathbf{x}_i)
\right|^2
\right)^{1/2}
},
\label{eq:test_relative_l2}
\end{equation}
where $N_q=3000$ denotes the number of sampled
spatial query points. Errors are averaged across
test cases, and the reported mean and sample
standard deviation are computed over the 10
independent training seeds. Each physical field
is evaluated separately.

GPU memory and runtime are measured directly during training. Peak allocated GPU memory is obtained using \texttt{torch.cuda.max\_memory\_allocated} after resetting the CUDA peak-memory statistics, and cold- and hot-chain step times are measured using \texttt{time.perf\_counter} with CUDA synchronization before and after the timed GPU operations. The logged per-step timing statistic is formed from the most recent 100 iterations. For the hot-phase resource comparison reported below, each seed is summarized using the median over the final 50\% of logged measurement windows, and the reported values are the mean $\pm$ sample standard deviation across 10 seeds. The memory quantity should therefore be interpreted as \emph{hot-phase peak allocated GPU memory with both replicas resident}, rather than as the whole-job system peak. Total training time, when reported separately, denotes end-to-end wall-clock time from the beginning of training. Throughout the replica-exchange experiments, the notation A/B denotes the precision policies of the cold and hot replicas, respectively.

\begin{table}[!htbp]
\centering
\small
\setlength{\tabcolsep}{5pt}
\renewcommand{\arraystretch}{1.10}
\caption{Default configuration for the primary replica-exchange experiments.
Unless otherwise stated, the same settings are used for the jet-engine
bracket and AirfRANS benchmarks.}
\label{tab:default_experimental_setup}
\begin{tabular}{@{}ll@{}}
\toprule
\textbf{Parameter} & \textbf{Value} \\
\midrule
Training iterations & 40,000 \\
Case batch size & 32 \\
Query-point batch size & 3,000 \\
Geometry point-cloud size & 3,000 \\
Transformer embedding dimension & 128 \\
Initial learning rate & $3\times10^{-4}$ \\
Learning-rate schedule & multiply by $0.99$ every 100 steps \\
Cold-chain temperature $T_c$ & $10^{-6}$ \\
Hot-chain temperature $T_h$ & $10^{-5}$ \\
Swap frequency & 100 iterations \\
reSGLD burn-in & 100 iterations \\
Evaluation frequency & 100 iterations \\
Independent seeds & 10 \\
AA-MP hot precision & BF16 live model; FP32 master/moment states \\
AMP hot precision & BF16 autocast with FP32 model parameters \\
\bottomrule
\end{tabular}
\end{table}

\subsection{Benchmark Problems}

\subsubsection{AirfRANS}
AirfRANS introduced in \cite{bonnet2022airfrans} is a high-fidelity computational fluid dynamic dataset for learning surrogate models of two-dimensional steady-state incompressible Reynolds-Averaged Navier--Stokes (RANS) solutions around airfoils in a subsonic regime. The dataset is designed for the aerodynamic shape-optimization setting, where the objective is to predict both the full flow fields and physically meaningful quantities such as lift and drag coefficients. Each simulation corresponds to an airfoil geometry, a Reynolds number, and an angle of attack (AOA).

The flow is modeled using the incompressible RANS equations. Denoting ensemble-averaged quantities by an overbar, the governing equations are written as
\begin{equation}
\partial_j(\bar{u}_i\bar{u}_j)
=
-\partial_i \bar{p}
+
\partial_j
\left[
(\nu+\nu_t)
\left(
\partial_j\bar{u}_i+\partial_i\bar{u}_j
\right)
\right],
\end{equation}
where $\bar{u}_i$ is the mean velocity field, $\bar{p}=p_{\mathrm{phys}}/\rho$ denotes the reduced (kinematic) pressure, $\rho$ is the fluid density, $\nu$ is the kinematic viscosity, and $\nu_t$ is the turbulent kinematic viscosity. The turbulent viscosity is modeled using the $k$--$\omega$ SST turbulence model.

The airfoils are generated from the NACA 4-digit and 5-digit families. The 4-digit family is parameterized by a sequence $MPXX$, where $M$ denotes the maximum ordinate of the camber line, $P$ its chordwise position, and $XX$ the maximum thickness. The 5-digit case is parameterized by $LPQXX$, where $L$ and $P$ define the camber, $Q$ distinguishes single- and double-cambered airfoils, and $XX$ again denotes the maximum thickness. The dataset contains $1000$ simulations. The Reynolds number ranges from $2\times 10^6$ to $6\times 10^6$, and the AOA ranges from $-5^\circ$ to $15^\circ$.

The simulations are generated using OpenFOAM with the steady-state RANS solver simpleFOAM, the SIMPLEC algorithm, and the $k$--$\omega$ SST turbulence model. Each airfoil is discretized using a C-grid mesh generated with \texttt{blockMesh}. The far-field boundary is placed at $200$ chord lengths from the airfoil to reduce boundary-condition effects. To accurately resolve surface quantities such as wall shear stress, the boundary-layer mesh is refined so that the first cells near the airfoil surface have height $2\,\mu$m, giving a worst-case $y^+$ of approximately $1$. The resulting meshes contain approximately $250{,}000$--$300{,}000$ cells.

The machine-learning task is to approximate the spatial fields
\begin{equation}
    \left(\bar{u}_x, \bar{u}_y, \bar{p}, \nu_t\right),
\end{equation}
where $\bar{u}_x$ and $\bar{u}_y$ are the mean velocity components, $\bar{p}$ is the reduced pressure, and $\nu_t$ is the turbulent kinematic viscosity.

The experiments use the processed dataset generated by the repository's conversion pipeline from all 1,000 simulations. The conversion used for the reported runs retains the full spatial simulation data rather than applying the rectangular crop used in an earlier preprocessing notebook. The spatial coordinates and signed-distance information are retained as geometric inputs, while the flow variables and freestream operating parameters are standardized using statistics computed during construction of the processed dataset. In the current processed file, these normalization statistics were computed over the complete 1,000-case collection before the fixed 800/200 train--test partition was used by the training loader. Thus, test simulations do not enter the optimization batches, but their distribution contributes to the preprocessing statistics. During training and evaluation, the loader subsamples query points from the processed simulation fields to control computational cost.

The present study uses the AirfRANS full-data regime, consisting of 800 training simulations and 200 test simulations. The same fixed partition is used for all precision configurations. We evaluate the four predicted fields $(\bar{u}_x,\bar{u}_y,\bar{p},\nu_t)$ and use AirfRANS as the
primary benchmark for examining predictive convergence, precision allocation, and computational efficiency under different numerical precision policies.

\subsubsection{Jet-Engine Bracket}

We use a three-dimensional jet-engine bracket dataset released with the Point-DeepONet study~\cite{Park_2026}, built from geometries derived from the DeepJEB benchmark~\cite{hong2024deepjeb3ddeeplearningbased}. The processed dataset used in this work contains 6,315 simulation cases corresponding to 2,105 unique bracket geometries, with three loading conditions (vertical, horizontal, and diagonal) associated with each geometry.
The predicted structural fields are the three displacement components $(u_x,u_y,u_z)$ and the von Mises stress $\sigma_{\mathrm{vm}}$.

The Point-DeepONet source study~\cite{Park_2026} reports nonlinear static finite-element analyses performed with Altair OptiStruct using second-order tetrahedral elements and a Ti--6Al--4V elastic--plastic material model with linear isotropic hardening. We refer to this source for the simulation setup; the processed data used here contain the resulting pointwise structural fields and operating parameters rather than the original finite-element solver decks.

The 6,315 simulation cases are partitioned into 5,052 training and 1,263 test cases using a fixed 80/20 case-level split. This partition is constructed over geometry--loading-condition pairs rather than unique geometries. Of the 1,022 unique geometries represented in the test set, 1,004 (98.2\%) also occur in training under another loading condition. Consequently, this benchmark measures generalization to held-out simulation cases and loading configurations, rather than generalization to unseen geometries. We use it to assess the consistency of the proposed precision strategy on a three-dimensional structural-mechanics problem.

\FloatBarrier

\subsection{Computational Efficiency}
\label{sec:computational_efficiency}

\begin{figure}[!htbp]
    \centering
    \includegraphics[width=\textwidth]
    {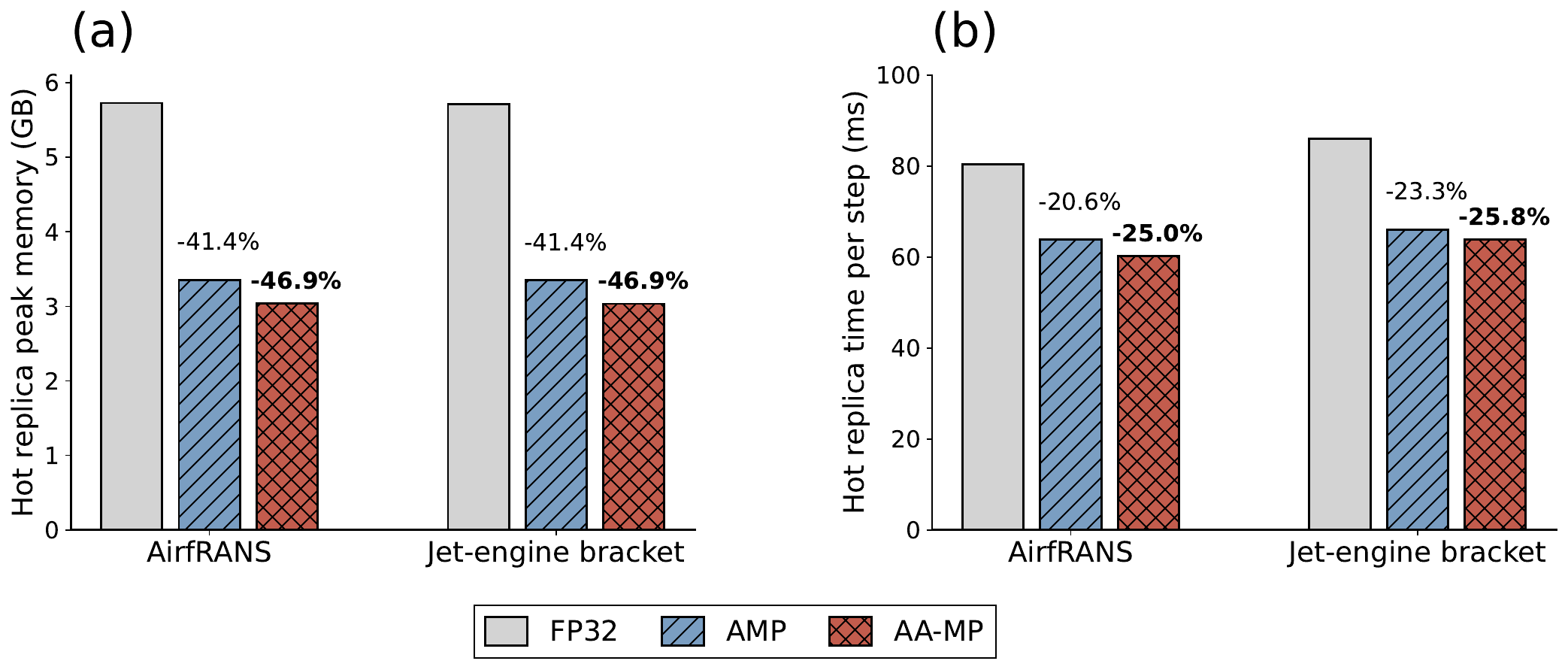}
    \caption{Hot-replica computational efficiency for AirfRANS and the jet-engine bracket benchmarks. (a) Hot-replica peak allocated GPU memory with both replicas resident. (b) Hot-replica execution time per training step. Bars show the mean over 10 seeds, with each seed summarized by the median over the final 50\% of logged measurement windows; error bars indicate sample standard deviation. Percentage annotations denote reductions relative to FP32/FP32.}
    \label{fig:hot_chain_efficiency}
\end{figure}

\begin{table}[!htbp]
\centering
\small
\setlength{\tabcolsep}{5pt}
\renewcommand{\arraystretch}{1.08}
\caption{
Hot-replica resource measurements corresponding to Fig.~\ref{fig:hot_chain_efficiency}.
Values are mean $\pm$ sample standard deviation across 10 seeds after per-seed
aggregation.
}
\label{tab:hot_chain_efficiency}
\begin{tabular}{@{}llcc@{}}
\toprule
\textbf{Dataset} & \textbf{Configuration}
& \textbf{Peak alloc (GB)}
& \textbf{Hot sec/step} \\
\midrule
\multirow{3}{*}{AirfRANS}
& FP32/FP32   & $5.727\pm0.000$ & $0.0804\pm0.0024$ \\
& FP32/AMP    & $3.356\pm0.000$ & $0.0638\pm0.0016$ \\
& FP32/AA-MP  & \bm{$3.042\pm0.000$} & \bm{$0.0603\pm0.0006$} \\
\midrule
\multirow{3}{*}{Bracket}
& FP32/FP32   & $5.709\pm0.000$ & $0.0861\pm0.0018$ \\
& FP32/AMP    & $3.347\pm0.000$ & $0.0660\pm0.0027$ \\
& FP32/AA-MP  & \bm{$3.034\pm0.000$} & \bm{$0.0639\pm0.0007$} \\
\bottomrule
\end{tabular}
\end{table}

Figure~\ref{fig:hot_chain_efficiency} summarizes the computational effect of reducing the precision of the exploratory replica. On AirfRANS, FP32/AA-MP reduces the hot-phase peak allocated memory from $5.727$ GB to $3.042$ GB, a $46.9\%$ reduction relative to FP32/FP32, while the hot-chain step time decreases from $0.0804$ s to $0.0603$ s, corresponding to a $25.0\%$ reduction. The FP32/AMP baseline also reduces the corresponding memory and step-time costs, to $3.356$ GB and $0.0638$ s per step, respectively.

The jet-engine bracket benchmark exhibits a similar pattern. FP32/AA-MP reduces hot-phase peak allocated memory from $5.709$ GB to $3.034$ GB and hot-chain time from $0.0861$ s to $0.0639$ s per step, corresponding to reductions of $46.9\%$ and $25.8\%$, respectively. FP32/AMP achieves $3.347$ GB and $0.0660$ s per hot-chain step. Across both benchmarks, AA-MP reduces hot-phase resource requirements relative to FP32/FP32 and provides additional savings over standard BF16 AMP. Relative to AMP, AA-MP reduces peak allocated GPU memory by approximately 9.4\% on both benchmarks, while hot-replica step time decreases by approximately 5.5\% on AirfRANS and 3.2\% on the bracket benchmark.

These measurements characterize the hot-chain phase rather than the whole-job system peak or end-to-end training speedup. Because the cold replica remains in FP32 and exchange evaluation, validation, data handling, and other training overheads are unaffected by the hot-replica precision policy, reductions in total wall-clock training time are necessarily smaller than the isolated hot-step reductions.

\subsection{Predictive Accuracy Across Precision Strategies}
\label{sec:predictive_accuracy}

\subsubsection{AirfRANS Benchmark}

\begin{figure}[!htbp]
    \centering
    \includegraphics[width=0.95\textwidth]{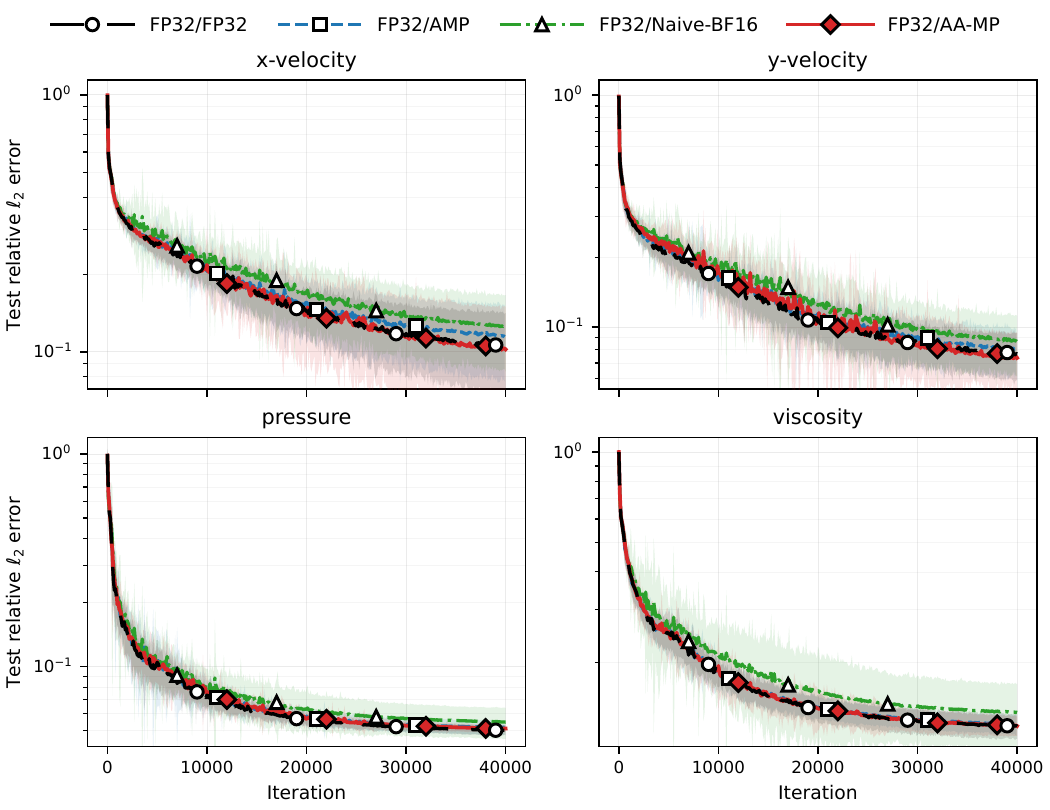}    
    \caption{AirfRANS reSGLD per-field test relative $\ell_2$ error. Curves show the mean over 10 independent seeds, while shaded regions indicate empirically calibrated pointwise bands based on across-seed variability.}
    \label{fig:airfrans_accuracy}
\end{figure}

AirfRANS provides the primary precision-sensitive benchmark in this study. Figure~\ref{fig:airfrans_accuracy} compares the cold-replica test relative $\ell_2$ error for FP32/FP32, FP32/AMP, FP32/Naive-BF16, and the proposed FP32/AA-MP configuration. The FP32/AA-MP trajectories remain close to the FP32/FP32 reference across the four predicted flow quantities, whereas
the less-protected Naive-BF16 configuration exhibits a clearer degradation, particularly for the velocity and turbulent-viscosity fields. Pressure is less sensitive to the tested precision policies. These results indicate that reduced precision can be applied to the exploratory replica while retaining predictive accuracy close to the FP32 reference when the BF16 live model is combined with FP32 master parameters, optimizer states, and DeepONet output and energy reductions.

\subsubsection{Jet-Engine Bracket Benchmark}

\begin{figure}[!htbp]
    \centering
    \includegraphics[width=0.95\textwidth]{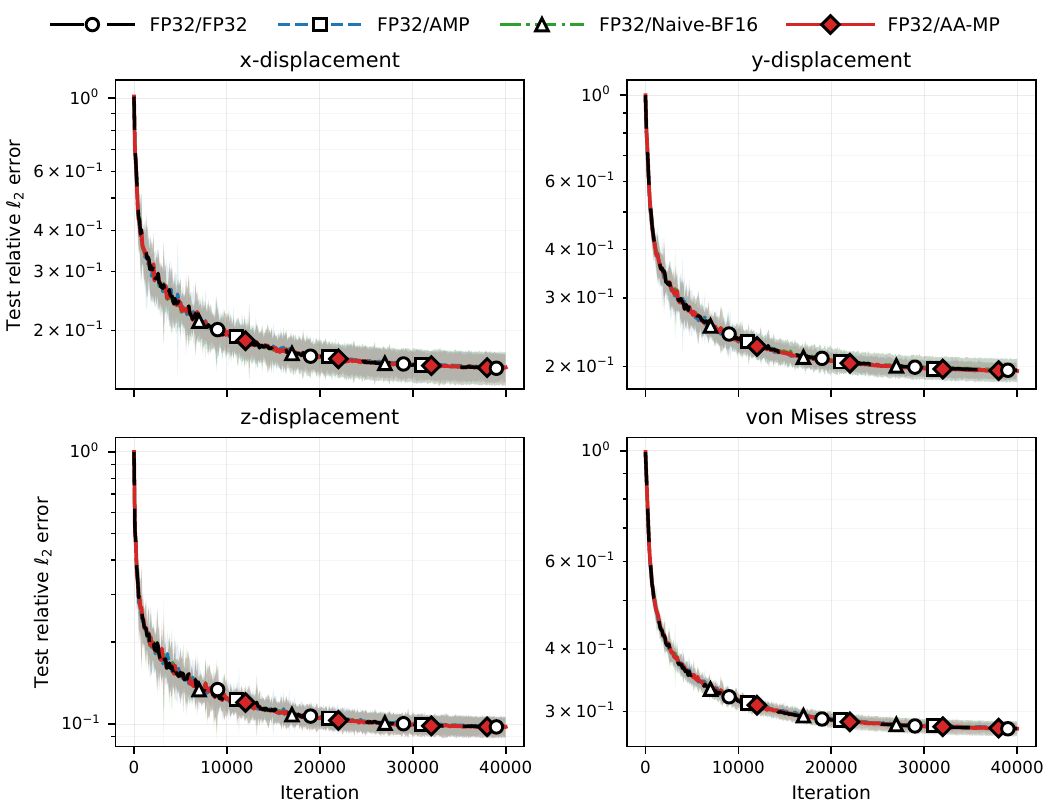}    
    \caption{Jet-engine bracket reSGLD cold-replica per-field test relative $\ell_2$ error. Curves show the mean over 10 independent seeds, while shaded regions indicate empirically calibrated pointwise bands based on across-seed variability.}
    \label{fig:bracket_accuracy}
\end{figure}

Figure~\ref{fig:bracket_accuracy} provides a complementary evaluation on
the jet-engine bracket benchmark. In contrast to AirfRANS, the four precision
strategies produce nearly overlapping convergence trajectories for the three
displacement components and von Mises stress. The bracket problem is therefore
less discriminative with respect to the tested precision policies, but it
provides evidence that the proposed strategy remains stable across a distinct
three-dimensional structural-mechanics problem.

To complement the convergence trajectories, Table~\ref{tab:final_accuracy} reports the per-field test relative $\ell_2$ errors aggregated over the same 10 seeds for each precision strategy. All four reSGLD precision configurations use the same model architecture, optimizer, learning-rate schedule, replica temperatures, batch sizes, data partition, and 40,000-update training budget; they differ only in the precision policy assigned to the hot replica.

\begin{table}[!htbp]
\centering
\caption{Per-field test relative $\ell_2$ errors across the evaluated precision strategies. For the AirfRANS and bracket reSGLD blocks, the cold replica remains in FP32 and the column header denotes the hot-replica precision policy. The bracket SGD block reports single-chain precision-policy controls in the corresponding column order; the cold/hot replica notation in the column headings applies only to the reSGLD results. Values are mean $\pm$ sample standard deviation over 10 seeds. Naive-BF16 is included as a reduced-precision ablation. Bold indicates the lowest mean error for each field.}
\label{tab:final_accuracy}
\begin{tabular}{@{}llcccc@{}}
\toprule
\textbf{Dataset} & \textbf{Field}
  & \textbf{FP32/FP32}
  & \textbf{FP32/AA-MP}
  & \textbf{FP32/AMP}
  & \textbf{FP32/Naive-BF16} \\
\midrule
\multirow{4}{*}{\shortstack[l]{AirfRANS\\(reSGLD)}}
  & $u$              & $0.1055 \pm 0.0155$ & $\mathbf{0.1022} \pm 0.0210$ & $0.1145 \pm 0.0158$ & $0.1250 \pm 0.0179$ \\
  & $v$              & $0.0765 \pm 0.0069$ & $\mathbf{0.0736} \pm 0.0078$ & $0.0795 \pm 0.0080$ & $0.0865 \pm 0.0104$ \\
  & $p$              & $\mathbf{0.0501} \pm 0.0011$ & $0.0511 \pm 0.0008$ & $0.0518 \pm 0.0014$ & $0.0548 \pm 0.0035$ \\
  & $\nu_t$          & $\mathbf{0.1226} \pm 0.0054$ & $0.1229 \pm 0.0032$ & $0.1244 \pm 0.0043$ & $0.1363 \pm 0.0102$ \\
\midrule
\multirow{4}{*}{\shortstack[l]{Bracket\\(reSGLD)}}
  & $u_x$            & $\mathbf{0.1543} \pm 0.0073$ & $0.1546 \pm 0.0069$ & $0.1553 \pm 0.0064$ & $0.1548 \pm 0.0075$ \\
  & $u_y$            & $0.1954 \pm 0.0055$ & $\mathbf{0.1945} \pm 0.0043$ & $0.1947 \pm 0.0041$ & $0.1959 \pm 0.0060$ \\
  & $u_z$            & $0.0977 \pm 0.0040$ & $\mathbf{0.0976} \pm 0.0037$ & $0.0984 \pm 0.0036$ & $0.0980 \pm 0.0036$ \\
  & $\sigma_{vm}$    & $\mathbf{0.2768} \pm 0.0046$ & $0.2770 \pm 0.0034$ & $0.2781 \pm 0.0027$ & $0.2777 \pm 0.0042$ \\
\midrule
\multirow{4}{*}{\shortstack[l]{Bracket\\(SGD)}}
  & $u_x$            & $0.3587 \pm 0.0229$ & $\mathbf{0.3554} \pm 0.0232$ & $0.3637 \pm 0.0330$ & $1.0046 \pm 0.0084$ \\
  & $u_y$            & $0.4729 \pm 0.0328$ & $0.4746 \pm 0.0363$ & $\mathbf{0.4714} \pm 0.0431$ & $1.0030 \pm 0.0054$ \\
  & $u_z$            & $0.2368 \pm 0.0194$ & $\mathbf{0.2306} \pm 0.0152$ & $0.2368 \pm 0.0259$ & $1.0032 \pm 0.0032$ \\
  & $\sigma_{vm}$    & $0.5520 \pm 0.0257$ & $\mathbf{0.5502} \pm 0.0261$ & $0.5529 \pm 0.0318$ & $0.9944 \pm 0.0021$ \\
\bottomrule
\end{tabular}
\end{table}

The per-field error statistics are consistent with the convergence trajectories in Figures~\ref{fig:airfrans_accuracy}
and~\ref{fig:bracket_accuracy}.
Across both benchmarks, FP32/AA-MP achieves predictive accuracy comparable to FP32/FP32 and
FP32/AMP. Although AA-MP attains lower mean errors for several individual fields, these differences are generally small relative to the variability across training seeds and are not interpreted as evidence of systematically improved predictive
accuracy. In contrast, Naive-BF16 exhibits larger errors for several AirfRANS quantities. These results support the use of replica-dependent
precision allocation to reduce computational resource requirements while maintaining comparable
predictive performance.

\FloatBarrier

\subsection{Qualitative Validation}
\label{sec:qualitative_fields}

Aggregate relative $\ell_2$ errors quantify predictive accuracy but do not show where approximation errors occur in the physical domain. We therefore examine representative solution fields to assess whether AA-MP preserves the dominant spatial structures of the learned solutions and to identify localized error patterns.

\subsubsection{AirfRANS}
\begin{figure}[H]
    \centering
    \includegraphics[width=0.85\textwidth]{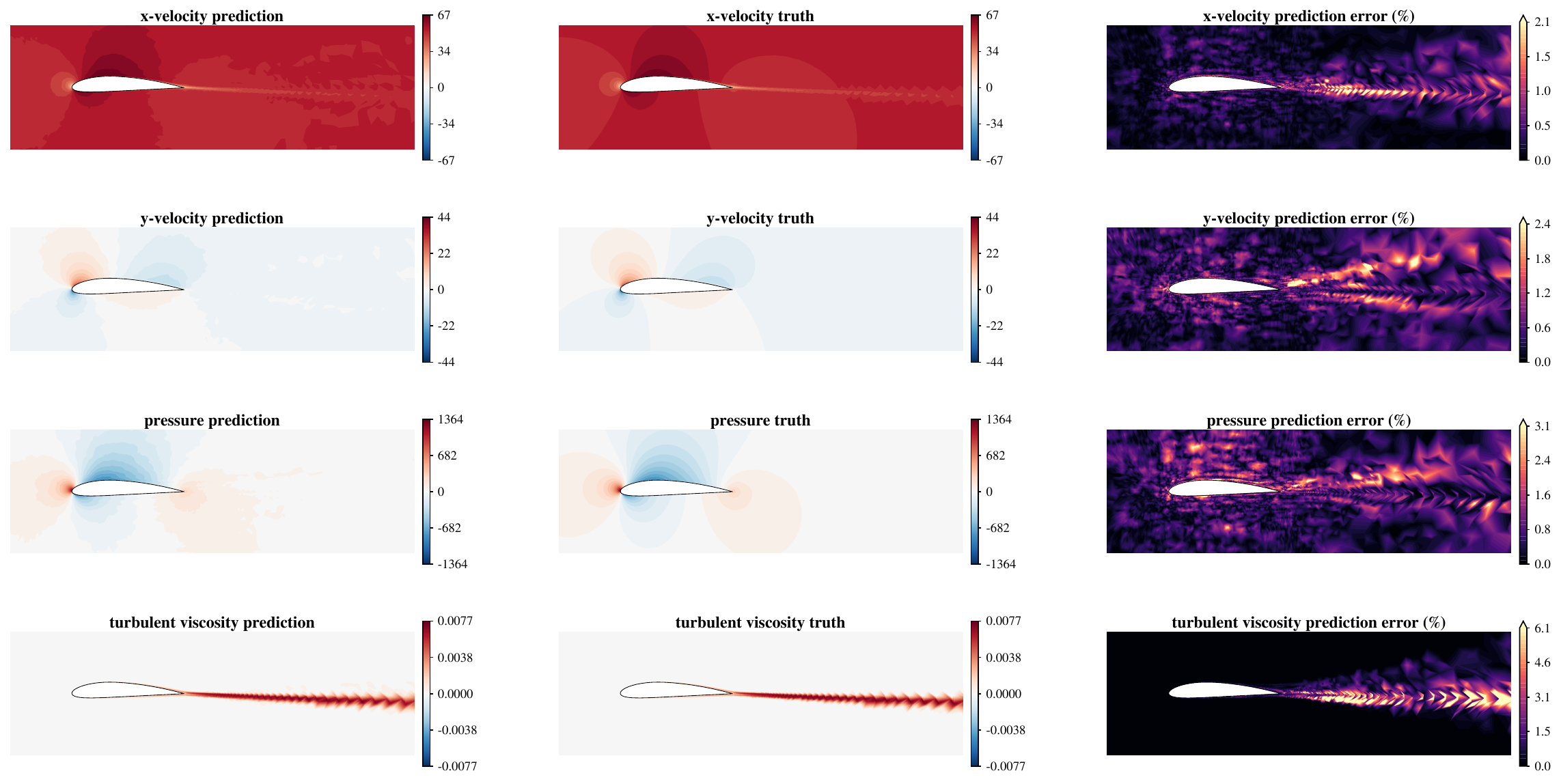}      
    \caption{Illustrative AirfRANS solution fields obtained with AA-MP. For each predicted quantity, the columns show the prediction, reference solution, and pointwise percentage error normalized by the maximum absolute reference value for that field. Prediction and reference fields use identical color limits within each field.}
    \label{fig:airfrans_sol_field_aamp}
\end{figure}

Figure~\ref{fig:airfrans_sol_field_aamp} compares the AA-MP prediction with the reference solution and the corresponding pointwise percentage error for the streamwise velocity, transverse velocity, pressure, and turbulent viscosity. For each field, the percentage error is defined as $100|\widehat{u}_k(\mathbf{x})-u_k(\mathbf{x})|/\max_j|u_k(\mathbf{x}_j)|$, where the maximum is taken over the full reference simulation case. The predicted fields reproduce the major flow structures surrounding the airfoil, while the error maps reveal the spatial regions in which the operator approximation is least accurate.

\subsubsection{Bracket}

\begin{figure}[H]
    \centering
    \includegraphics[width=0.80\textwidth]{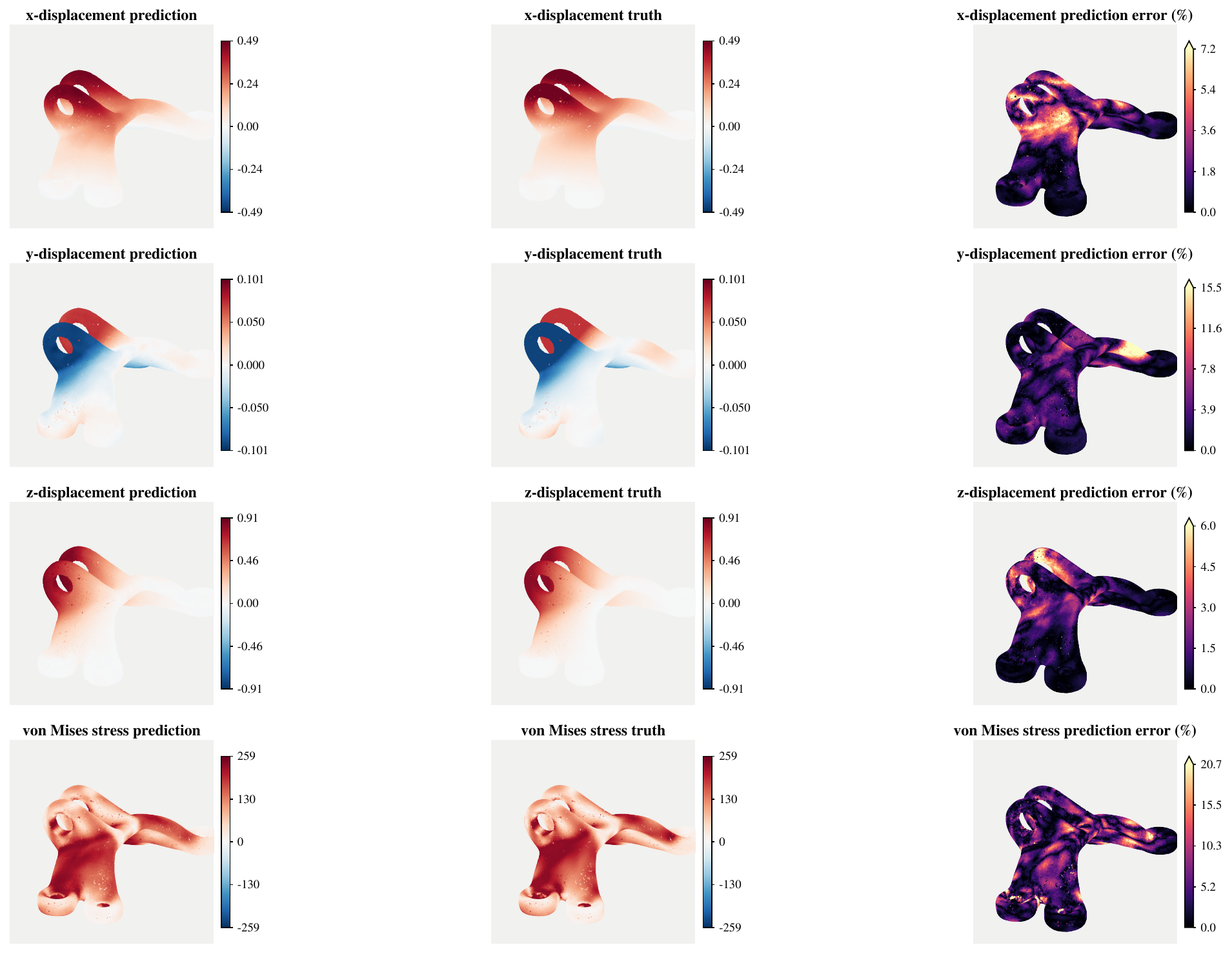}       
    \caption{Illustrative jet-engine bracket solution fields obtained with AA-MP. For each predicted quantity, the columns show the prediction, reference solution, and pointwise percentage error normalized by the maximum absolute reference value for that field. Prediction and reference fields use identical color limits within each field.}
    \label{fig:bracket_sol_field_aamp}
\end{figure}

\FloatBarrier

\subsection{Architecture-Level FP32 Protection Ablation}
\label{sec:architecture_precision}

To examine whether retaining additional model components in FP32 improves predictive performance, we performed a progressive precision-protection ablation on the AirfRANS benchmark. Starting from the proposed AA-MP policy, FP32 computation was successively introduced into the coordinate/RoPE pathway, K/V normalization, the input-feature pathway, and the Q/K/V projections. All configurations use the same 20,000-iteration screening protocol and 10 paired seeds.

\begin{table}[t]
\centering
\small
\caption{Per-field predictive accuracy for the architecture-level FP32-protection ablation on AirfRANS. All configurations use the same 20,000-iteration training protocol and 10 paired seeds. Values are mean per-case test relative $\ell_2$ error, reported as mean $\pm$ sample standard deviation across seeds. The FP32-protection rows are cumulative in their displayed order: RoPE, K/V LayerNorm, the input-feature pathway, and Q/K/V projections. Each successive row retains the protections of the preceding row. Bold indicates the lowest mean error for each field.}
\label{tab:precision_sensitivity_accuracy}
\begin{tabular}{lcccc}
\toprule
\textbf{Precision policy}
& $u$
& $v$
& $p$
& $\nu_t$ \\
\midrule

\textbf{AA-MP}
& $0.0593 \pm 0.0048$
& $0.0684 \pm 0.0046$
& $0.0524 \pm 0.0016$
& $0.0988 \pm 0.0032$ \\

AA-MP + FP32 RoPE
& $0.0582 \pm 0.0062$
& $\mathbf{0.0642} \pm 0.0056$
& $\mathbf{0.0500} \pm 0.0018$
& $\mathbf{0.0936} \pm 0.0026$ \\

AA-MP + FP32 LayerNorm
& $0.0596 \pm 0.0072$
& $0.0674 \pm 0.0046$
& $0.0512 \pm 0.0012$
& $0.0953 \pm 0.0032$ \\

AA-MP + FP32 input path
& $0.0598 \pm 0.0085$
& $0.0668 \pm 0.0064$
& $0.0510 \pm 0.0025$
& $0.0941 \pm 0.0032$ \\

AA-MP + FP32 projections
& $\mathbf{0.0581} \pm 0.0056$
& $0.0669 \pm 0.0055$
& $0.0508 \pm 0.0020$
& $0.0959 \pm 0.0030$ \\

FP32/FP32 control
& $0.0609 \pm 0.0094$
& $0.0681 \pm 0.0089$
& $0.0516 \pm 0.0017$
& $0.0952 \pm 0.0029$ \\

\bottomrule
\end{tabular}
\end{table}

\begin{table}[!htbp]
\centering
\small
\setlength{\tabcolsep}{5pt}
\renewcommand{\arraystretch}{1.08}
\caption{
Hot-replica computational cost for the AirfRANS FP32-protection ablation. For each seed, resource measurements are summarized by the median over the final 50\% of logged measurement windows, using the same aggregation protocol as Table~\ref{tab:hot_chain_efficiency}. Values are then averaged over 10 seeds. The FP32 protections are cumulative in the same order as Table~\ref{tab:precision_sensitivity_accuracy}. Memory and step-time reductions are relative to the FP32/FP32 control.}
\label{tab:precision_sensitivity_efficiency}
\begin{tabular}{@{}lcccc@{}}
\toprule
\textbf{Precision policy}
& \textbf{Peak alloc. (GB)}
& \textbf{Mem. red.}
& \textbf{Hot sec/step}
& \textbf{Time red.} \\
\midrule

\textbf{AA-MP}
& \textbf{3.051}
& \textbf{46.8\%}
& \textbf{0.0580}
& \textbf{25.5\%} \\

AA-MP + FP32 RoPE
& 3.785
& 34.0\%
& 0.0655
& 15.8\% \\

AA-MP + FP32 LayerNorm
& 4.153
& 27.6\%
& 0.0685
& 12.0\% \\

AA-MP + FP32 input path
& 4.151
& 27.6\%
& 0.0690
& 11.4\% \\

AA-MP + FP32 projections
& 4.357
& 24.0\%
& 0.0728
& 6.5\% \\

FP32/FP32 control
& 5.734
& --
& 0.0778
& -- \\

\bottomrule
\end{tabular}
\end{table}

Table~\ref{tab:precision_sensitivity_accuracy} reveals a modest accuracy--efficiency tradeoff. Relative to AA-MP, retaining the RoPE pathway in FP32 reduces the mean test relative $\ell_2$ error by 6.2\% for $v$, 4.6\% for $p$, and 5.2\% for $\nu_t$. Paired tests across the 10 seeds show that these reductions remain statistically detectable after Holm correction ($p=0.015$, $0.002$, and $0.015$, respectively). The corresponding change in $u$ is smaller ($-1.9\%$) and is not
statistically detectable ($p=0.409$).

The accuracy improvement is accompanied by a substantial computational cost. As shown in Table~\ref{tab:precision_sensitivity_efficiency}, AA-MP reduces hot-replica peak allocated memory and hot-step time by 46.8\% and 25.5\%, respectively, relative to the FP32/FP32 control. Protecting the RoPE pathway reduces these savings to 34.0\% and 15.8\%. Equivalently, FP32 RoPE requires approximately 24\% more hot-replica memory and 13\% more hot-step time than AA-MP.

The additional protection stages do not yield a monotonic improvement in predictive accuracy. Among the protected configurations, FP32 RoPE gives the lowest mean errors for $v$, $p$, and $\nu_t$, while further protecting LayerNorm, the input pathway, and the Q/K/V projections progressively reduces the computational advantage. At the most protected setting, the memory and step-time savings decrease to 24.0\% and 6.5\%, respectively.

We therefore retain AA-MP as the default configuration for the primary experiments because it provides the largest computational savings among the evaluated mixed-precision policies while maintaining competitive predictive accuracy. The FP32-RoPE configuration represents an alternative, more accuracy-oriented operating point when modest improvements in $v$, $p$, or $\nu_t$ are prioritized.

The resource values in this ablation are obtained from the independent 20,000-iteration screening campaign and therefore differ slightly from the primary 40,000-iteration measurements in Table~\ref{tab:hot_chain_efficiency}; both studies use the same per-seed resource-aggregation procedure.

\subsection{Uncertainty Quantification Evaluation}
\label{sec:uq_evaluation}

We evaluate spatially resolved uncertainty quantification using cold-replica states obtained through a separate post-training Langevin sampling phase. Following 100,000 training iterations, each precision configuration undergoes 5,000 additional preconditioned reSGLD updates, yielding 50 retained cold-temperature states. These states provide predictive means and spatially varying uncertainty estimates, which are subsequently calibrated using conformal risk control. Conformal uncertainty quantification has also been investigated for DeepONet models~\cite{moya2025conformalizeddeeponet}. 

Predictions from the retained cold-replica states vary across spatial locations and model parameter states. Their empirical dispersion provides a spatially varying uncertainty scale associated with the learned operator and the post-training Langevin trajectories. This dispersion is subsequently adjusted through conformal calibration using observed prediction errors. The resulting prediction intervals therefore incorporate empirical predictive-error information in addition to the variability among retained states.

For field $k\in\{u,v,p,\nu_t\}$ and spatial point $\mathbf{x}$, let $\mu_k(\mathbf{x})$ and $s_k(\mathbf{x})$ denote the mean and standard deviation of the retained-state predictions. We construct the symmetric conformal prediction interval
\begin{equation}
\label{eq:uq_conformal_interval}
C_k(\mathbf{x})=
\left[\mu_k(\mathbf{x})-H_k(\mathbf{x}),\;\mu_k(\mathbf{x})+H_k(\mathbf{x})\right],
\qquad
H_k(\mathbf{x})=\widehat\lambda_k\sqrt{s_k^2(\mathbf{x})+\epsilon_k^2}.
\end{equation}

For spatial visualization and interval-sharpness comparison, we report the full conformal prediction-interval width,
\begin{equation}
\label{eq:uq_interval_width}
W_k(\mathbf{x})
= 2H_k(\mathbf{x})
= 2\widehat{\lambda}_k
\sqrt{s_k^2(\mathbf{x})+\epsilon_k^2}.
\end{equation}

The scale floor $\epsilon_k$ is estimated from 40 separate training-side cases, while the multiplier $\widehat\lambda_k$ is selected using conformal risk control (CRC)~\cite{angelopoulos2024conformalrisk} on 100 calibration cases. All four methods are evaluated on the same 98 held-out AirfRANS cases, with identical spatial coordinates and reference fields. CRC targets a nominal 5\% mean spatial miscoverage across exchangeable AirfRANS cases, corresponding to a case-averaged coverage target of 95\%. Interval construction uses consistent standardized output units, and the resulting intervals are converted to physical-field units before reporting widths and interval scores.

For spatial visualization, we select AirfRANS evaluation case~911 based on its proximity to the median physical characteristics within the majority airfoil-parameterization family, considering freestream velocity, angle of attack, thickness, and camber. The selection is independent of prediction errors and conformal coverage, and the same evaluation case is used for all four precision configurations.

\begin{figure}[!htbp]
    \centering
    \includegraphics[width=0.95\textwidth]{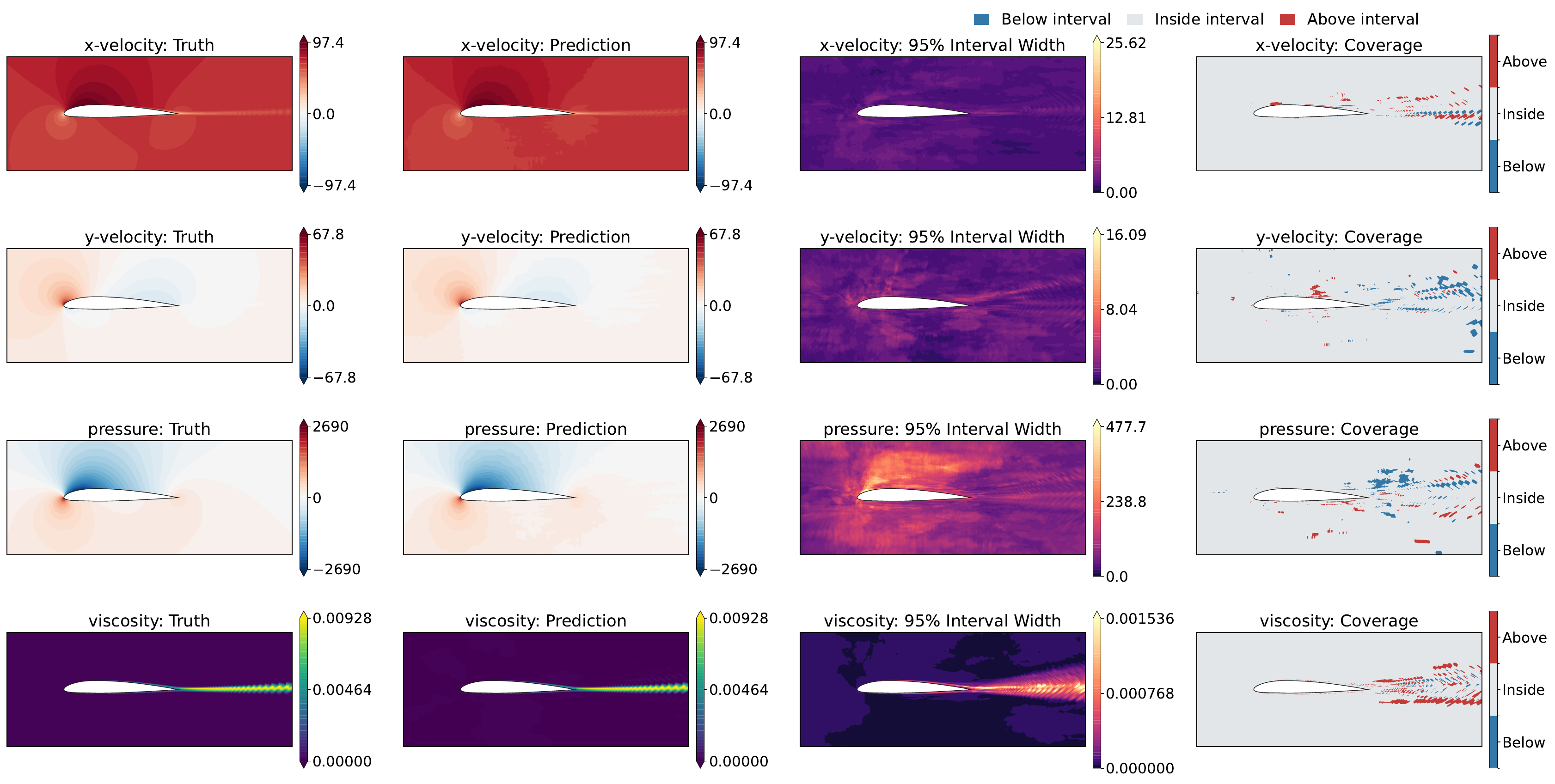}
    \caption{Spatial conformal uncertainty quantification for FP32/AA-MP on AirfRANS evaluation case~911. Rows show x-velocity $u$, y-velocity $v$, reduced pressure $p$, and turbulent viscosity $\nu_t$. Columns show the reference solution, predictive mean, full conformal-calibrated prediction-interval width $W_k=2H_k$, and spatial interval-coverage diagnostic. Gray indicates locations covered by the interval, red indicates reference values above its upper bound, and blue indicates values below its lower bound.}
    \label{fig:uq_aamp}
\end{figure}

Figure~\ref{fig:uq_aamp} illustrates the calibrated FP32/AA-MP intervals for case~911. For each flow field, the panels show the reference solution, predictive mean, full conformal prediction-interval width, and spatial coverage diagnostic. Corresponding visualizations for the other three precision configurations are provided in Appendix~\ref{app:uq_visualization} (Figs.~\ref{fig:uq_fp32}--\ref{fig:uq_naive_bf16}). These maps complement the aggregate results in Table~\ref{tab:uq_conformal_comparison}.

\begin{table}[!tbp]
\centering
\scriptsize
\setlength{\tabcolsep}{3.0pt}
\renewcommand{\arraystretch}{1.15}
\caption{Post-training AirfRANS conformal-UQ results for four precision configurations at nominal 95\% case-averaged node coverage (98 evaluation cases). Coverage and 5th-percentile per-case coverage are percentages; mean full prediction-interval width and interval score use de-normalized dataset output units. Smaller interval widths indicate sharper predictions, while lower interval scores indicate a more favorable combined sharpness--coverage trade-off. Bold indicates coverage closest to the nominal target, the smallest interval width and score, or the highest 5th-percentile coverage, as appropriate.}
\label{tab:uq_conformal_comparison}
\begin{tabular}{@{}llrrrr@{}}
\toprule
\textbf{Field} & \textbf{Method} & \textbf{Mean cov.} & \shortstack{\textbf{Mean full}\\\textbf{width}} & \textbf{Interval score} & \textbf{5th pct.} \\
\midrule
\multirow{4}{*}{$u$}
 & FP32/FP32       & 96.24 & 14.570 & 16.880 & 91.73 \\
 & FP32/AMP        & 95.94 & 23.300 & 26.400 & 91.59 \\
 & FP32/AA-MP      & \textbf{94.92} & \textbf{3.328} & \textbf{4.593} & 82.36 \\
 & FP32/Naive-BF16 & 95.95 & 28.564 & 31.537 & \textbf{93.45} \\
\midrule
\multirow{4}{*}{$v$}
 & FP32/FP32       & 96.24 & 14.370 & 16.690 & \textbf{92.80} \\
 & FP32/AMP        & 96.16 & 28.600 & 30.030 & 89.25 \\
 & FP32/AA-MP      & \textbf{95.08} & \textbf{2.940} & \textbf{3.861} & 81.91 \\
 & FP32/Naive-BF16 & 95.87 & 25.920 & 28.478 & 91.69 \\
\midrule
\multirow{4}{*}{$p$}
 & FP32/FP32       & 96.64 & 902.0  & 1155.0 & \textbf{92.52} \\
 & FP32/AMP        & 96.17 & 2530.0 & 2634.0 & 92.09 \\
 & FP32/AA-MP      & \textbf{95.24} & \textbf{190.6} & \textbf{292.3} & 83.48 \\
 & FP32/Naive-BF16 & 96.08 & 2140.2 & 2338.6 & 91.93 \\
\midrule
\multirow{4}{*}{$\nu_t$}
 & FP32/FP32       & 95.97 & 0.0004360 & 0.0012770 & \textbf{92.66} \\
 & FP32/AMP        & 95.54 & 0.0006592 & 0.0015200 & 92.59 \\
 & FP32/AA-MP      & \textbf{95.48} & \textbf{0.0002104} & \textbf{0.0007496} & 88.42 \\
 & FP32/Naive-BF16 & 96.12 & 0.0009412 & 0.0014996 & 91.01 \\
\bottomrule
\end{tabular}
\end{table}

\begin{figure}[!htbp]
    \centering
    \includegraphics[width=0.95\textwidth]{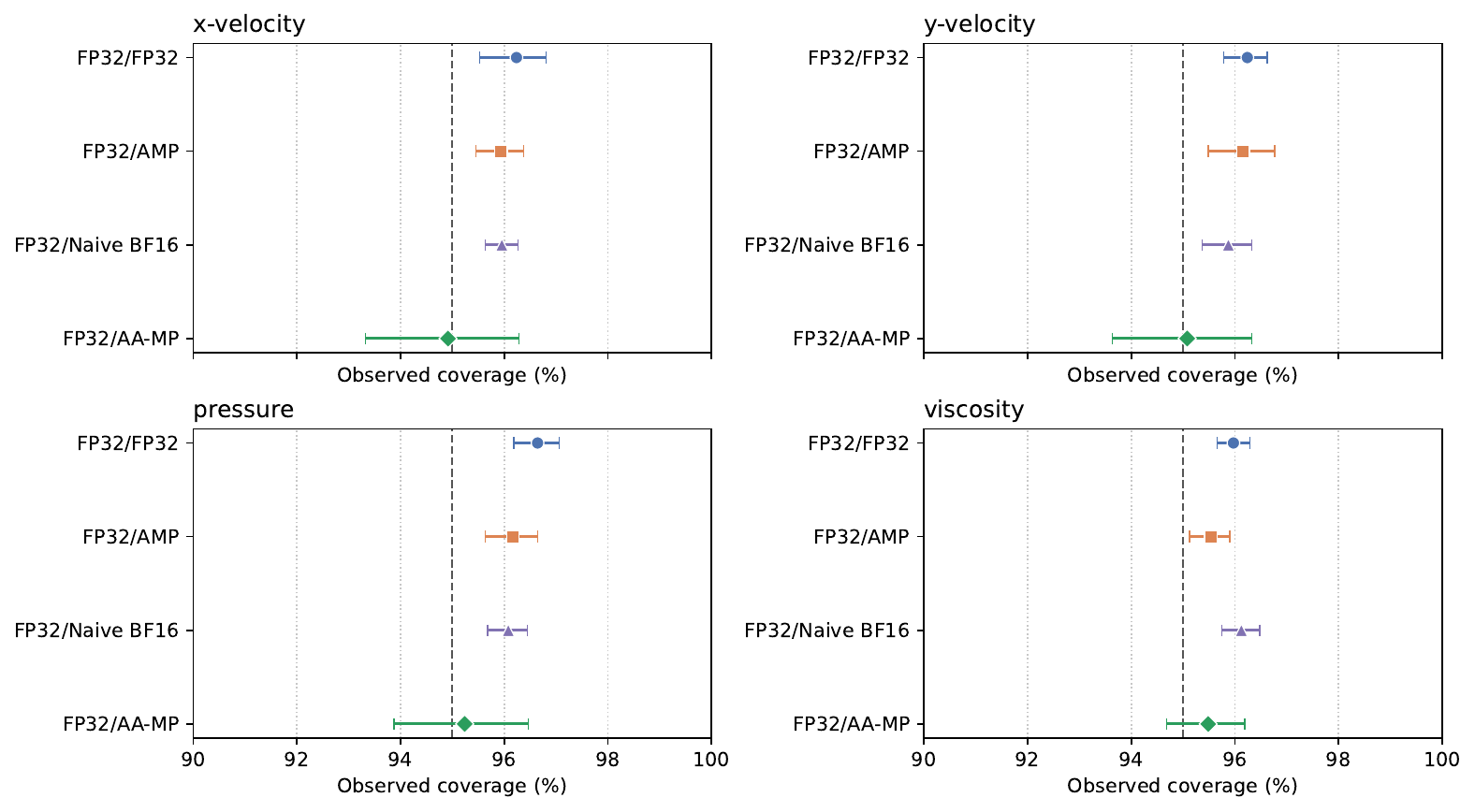}
    \caption{Empirical coverage of conformal prediction intervals across AirfRANS flow fields. Markers indicate the mean fraction of covered spatial nodes across 98 held-out CFD cases for each precision configuration. Horizontal error bars represent 95\% bootstrap confidence intervals obtained by resampling evaluation cases. The dashed vertical line denotes the nominal 95\% coverage target. All four configurations achieve mean coverage near the target, with FP32/AA-MP exhibiting coverage closest to 95\% across the predicted fields.}
    \label{fig:coverage_comparison}
\end{figure}

To evaluate interval sharpness together with miscoverage, we report the mean interval score~\cite{gneiting2007proper}. Let $y_k(\mathbf{x})$ denote the corresponding truth value for field $k$. For the symmetric interval in Eq.~\eqref{eq:uq_conformal_interval}, its pointwise value at nominal miscoverage $\alpha=0.05$ is
\begin{equation}
\label{eq:uq_interval_score}
S_{0.05,k}(\mathbf{x})
=
W_k(\mathbf{x})
+
40\left(
|y_k(\mathbf{x})-\mu_k(\mathbf{x})|
-\frac{W_k(\mathbf{x})}{2}
\right)_+,
\end{equation}
where $(a)_+=\max(a,0)$. The scores are averaged over nodes within each case and then equally over evaluation cases. Smaller scores indicate narrower intervals with an appropriate penalty for observations outside the interval.

As shown in Fig.~\ref{fig:coverage_comparison}, all four precision configurations achieve mean case-averaged spatial coverage near the nominal 95\% target. FP32/AA-MP exhibits coverage closest to the target across all four fields. Table~\ref{tab:uq_conformal_comparison} further shows that AA-MP achieves the smallest mean full prediction-interval width and lowest mean interval score across the evaluated fields. However, its lower 5th-percentile per-case coverage indicates greater variability in coverage among individual flow configurations.

Figure~\ref{fig:uq_aamp} provides a spatial illustration for case~911. The coverage maps distinguish locations where the reference field lies inside the prediction interval from those where it falls above or below the interval bounds. The observed violations are localized, including regions around and downstream of the airfoil. The nominal coverage target concerns case-averaged spatial coverage, not uniform coverage at every node or for every individual case. Corresponding diagnostics for the other precision configurations appear in Figs.~\ref{fig:uq_fp32},~\ref{fig:uq_amp}, and~\ref{fig:uq_naive_bf16}.

The retained cold-state dispersion is interpreted as an empirical measure of parameter-dependent uncertainty, as the finite post-training Langevin trajectories have not been independently verified for posterior mixing.

\subsection{Precision Allocation Across Replica Roles}
\label{sec:precision_allocation}
We study whether reduced precision should be assigned to the cold replica, the hot replica, or both.

\begin{table}[!htbp]
\centering
\small
\setlength{\tabcolsep}{5pt}
\renewcommand{\arraystretch}{1.12}
\caption{Ablation configurations used to evaluate replica-dependent precision allocation. All preconditioned reSGLD configurations use identical hyperparameters.}
\label{tab:replica_exchange_ablation}

\begin{tabularx}{\textwidth}{
@{}
>{\raggedright\arraybackslash}p{0.29\textwidth}
>{\raggedright\arraybackslash}X
@{}
}
\toprule
\textbf{Training configuration} & \textbf{Purpose} \\
\midrule

FP32/FP32
& Full-precision replica-exchange baseline used to quantify the effect of low precision. \\

\addlinespace
AA-MP/AA-MP
& Evaluates the effect of applying reduced precision to both the cold and hot chain. \\

\addlinespace
AA-MP/FP32
& Evaluates the effect of applying low precision to the cold chain while retaining full precision in the hot chain. \\

\addlinespace
FP32/AA-MP
& Proposed replica-dependent precision configuration, retaining full precision for cold-chain refinement while using low precision for hot-chain exploration. \\

\bottomrule
\end{tabularx}
\end{table}

\subsubsection{AirfRANS}
\begin{figure}[htbp!]
    \centering
    \includegraphics[width=0.90\textwidth]{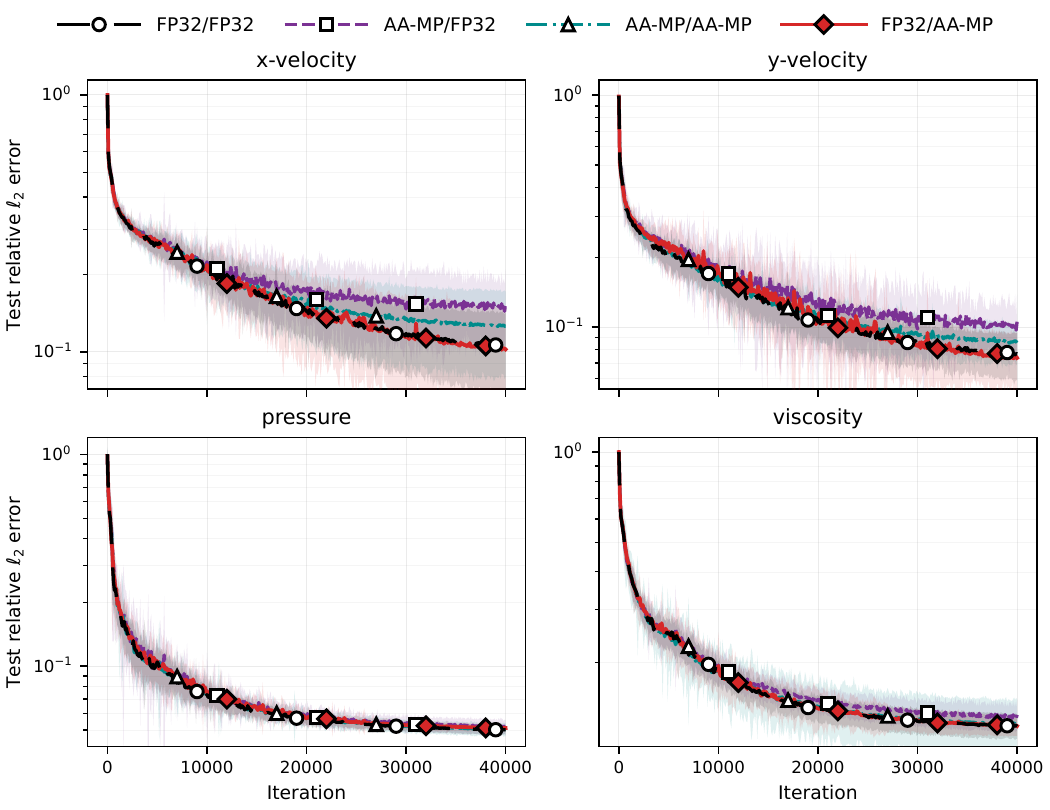} 
    \caption{AirfRANS cold-replica test relative $\ell_2$ error for four cold/hot precision allocations: FP32/FP32, AA-MP/FP32, AA-MP/AA-MP, and FP32/AA-MP. Curves show the mean over 10 independent seeds, while shaded regions indicate empirically calibrated pointwise bands based on across-seed variability. Configurations are denoted as cold-replica precision / hot-replica precision.}
    \label{fig:airfrans_allocation}
\end{figure}

AirfRANS provides the clearest evidence of precision-allocation sensitivity. Figure~\ref{fig:airfrans_allocation} shows that the FP32/AA-MP configuration closely follows the fully FP32 reference, whereas configurations that apply AA-MP to the cold replica exhibit a larger separation for the velocity and turbulent-viscosity fields. The pressure prediction is less sensitive to precision placement. This asymmetry supports retaining FP32 in the low-temperature refinement replica while assigning reduced precision to the higher-temperature exploratory replica.

\subsubsection{Jet-Engine Bracket}
\begin{figure}[htbp!]
    \centering
    \includegraphics[width=0.90\textwidth]{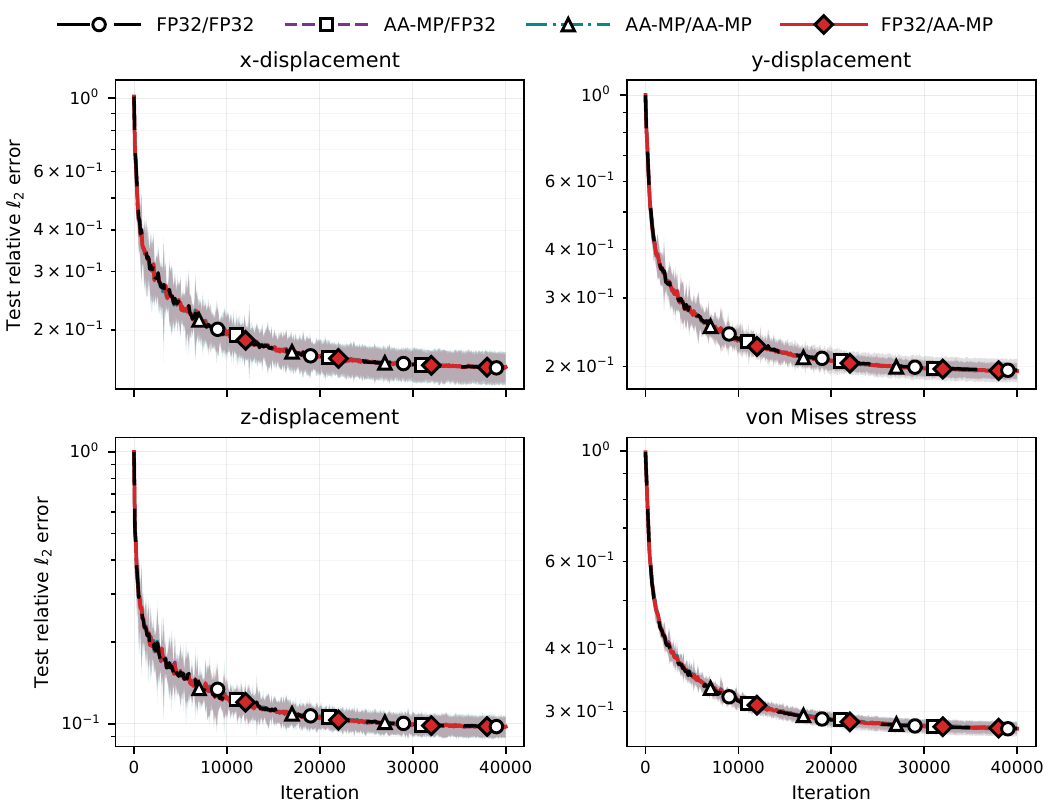}  
    \caption{Jet-engine bracket cold-replica test relative $\ell_2$ error for four cold/hot precision allocations: FP32/FP32, AA-MP/FP32, AA-MP/AA-MP, and FP32/AA-MP. Curves show the mean over 10 independent seeds, while shaded regions indicate empirically calibrated pointwise bands based on across-seed variability. Configurations are denoted as cold-replica precision / hot-replica precision.}
    \label{fig:bracket_allocation}
\end{figure}

Figure~\ref{fig:bracket_allocation} shows the corresponding result for the bracket benchmark. All four precision allocations produce nearly overlapping convergence trajectories for the reported structural fields. Although this benchmark does not strongly discriminate among the tested allocations, it confirms that FP32/AA-MP does not introduce a visible convergence penalty in this structural-mechanics problem.

\FloatBarrier

\paragraph{Replica-exchange activity}
Replica exchange remained active under both FP32/FP32 and FP32/AA-MP. Across 10 seeds, the mean swap-acceptance rate changed from $37.5\pm10.4\%$ for FP32/FP32 to $18.4\pm2.8\%$ for FP32/AA-MP on AirfRANS, and from $33.4\pm15.0\%$ to $29.0\pm16.6\%$ on the jet-engine bracket benchmark. These statistics are included only to confirm that exchange events remain active during training; the acceptance rate itself is not interpreted as a measure of predictive performance or sampling quality.

\subsection{Computational Scaling}

To examine how the computational benefit of AA-MP changes with geometry-processing workload, we vary the geometry point-cloud size while holding the remaining training configuration fixed. FP32/FP32, FP32/AMP, FP32/Naive-BF16, and FP32/AA-MP are compared at the same tested point-cloud sizes using the same reported resource metrics. Because the FP32 reference and reduced-precision measurements originate from different scaling workflows, these experiments are used to compare resource-scaling trends rather than absolute speedups.

\begin{figure}[!htbp]
    \centering
    \includegraphics[width=0.95\textwidth]{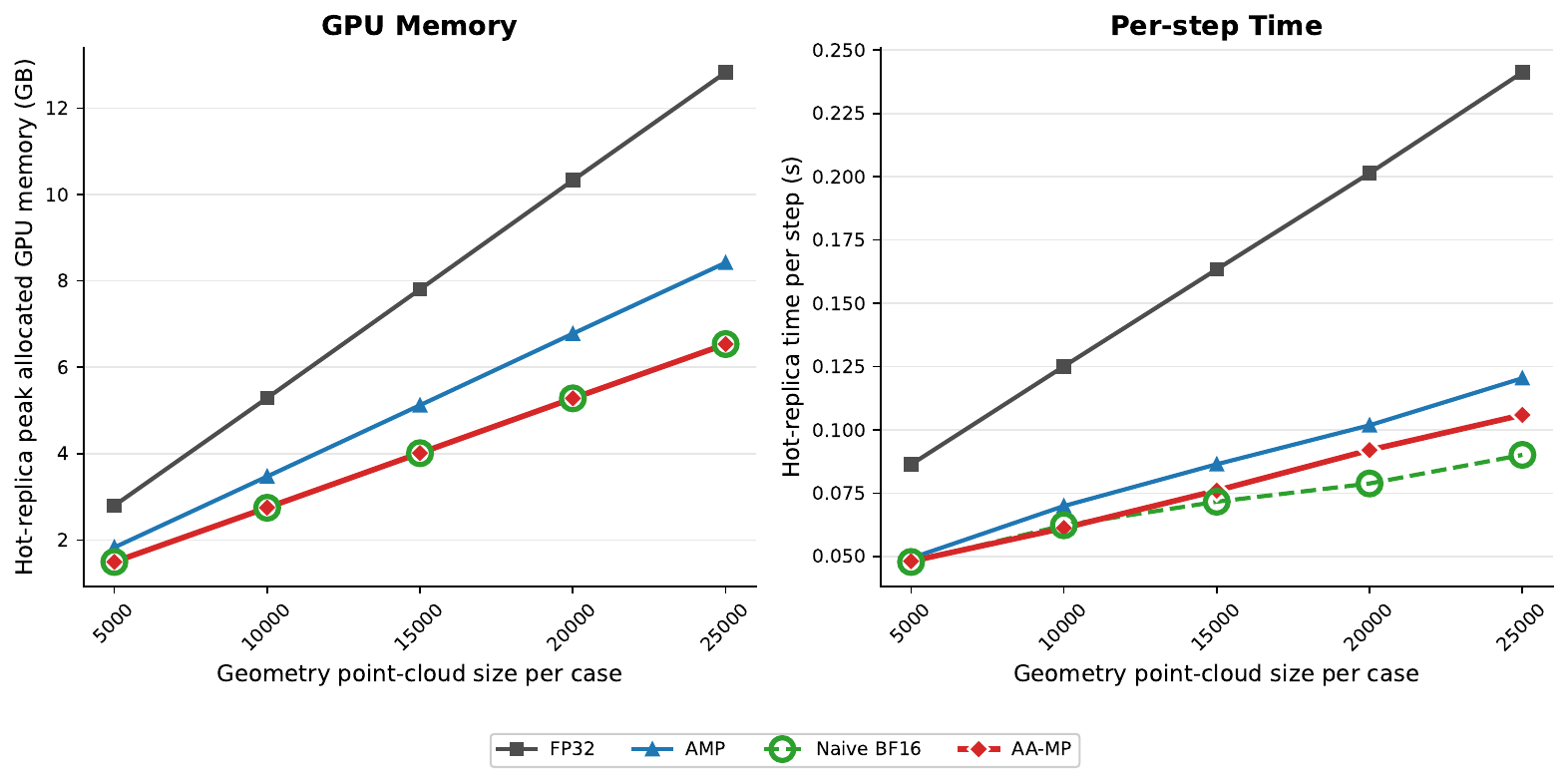}
    \caption{Geometry point-cloud-size resource scaling for FP32/FP32, FP32/AMP, FP32/Naive-BF16, and FP32/AA-MP. The geometry point-cloud size is varied from 5,000 to 25,000.}
    \label{fig:pointcloud_ablation}
\end{figure}

Figure~\ref{fig:pointcloud_ablation} shows GPU-memory and hot-replica step-time scaling as the geometry point-cloud size increases. AA-MP requires less allocated GPU memory than both FP32/FP32 and standard BF16 AMP across the tested sizes, with a widening absolute memory gap at larger point-cloud sizes. AA-MP also reduces the measured hot-replica step time relative to AMP across these configurations.

\begin{figure}[!htbp]
    \centering
    \includegraphics[width=0.95\textwidth]{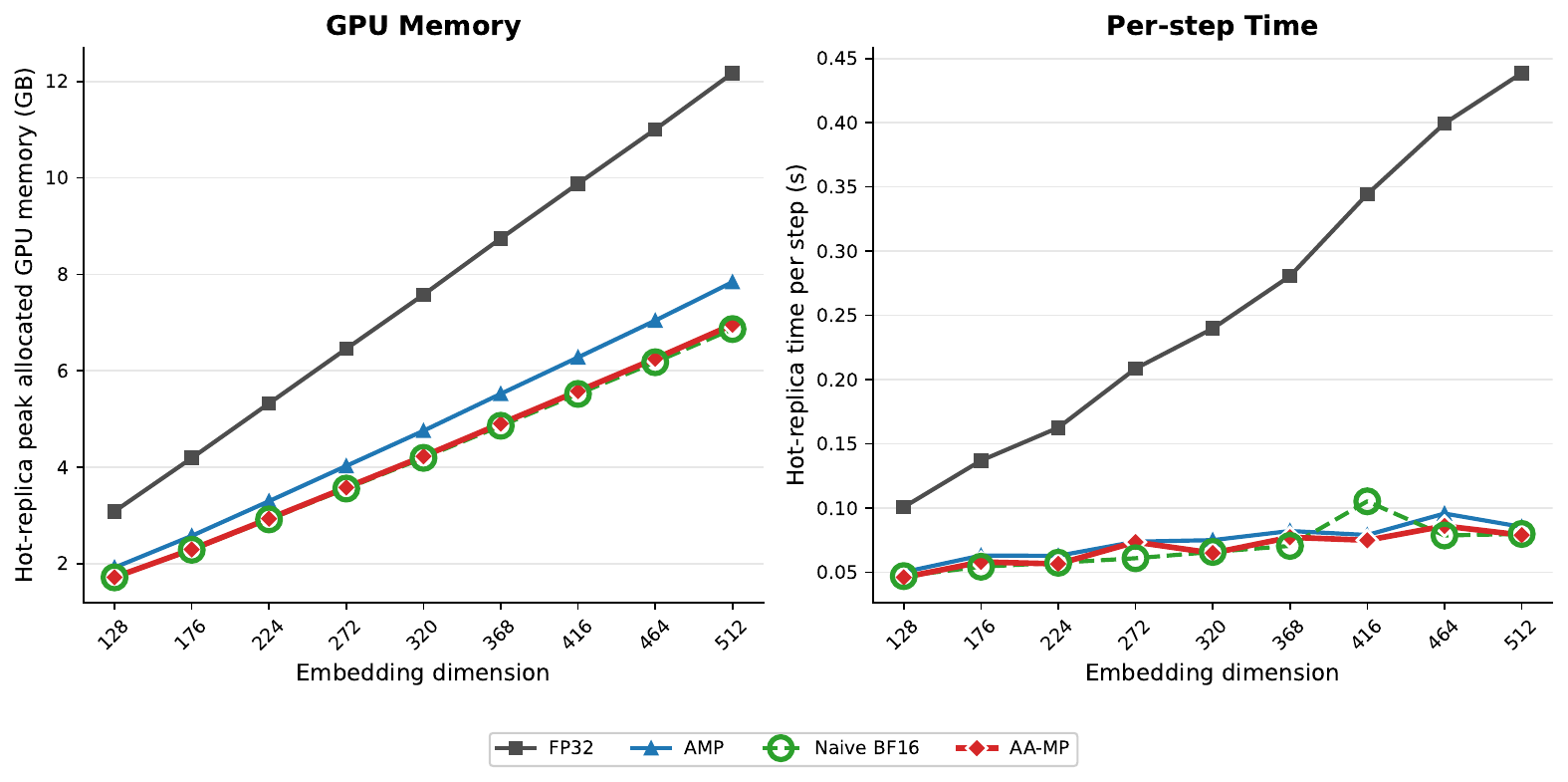}
    \caption{Embedding-dimension resource scaling for FP32/FP32, FP32/AMP, FP32/Naive-BF16, and FP32/AA-MP. The Transformer embedding dimension is varied from 128 to 512.}
    \label{fig:embedding_dim_ablation}
\end{figure}

Figure~\ref{fig:embedding_dim_ablation} shows resource scaling as the Transformer embedding dimension increases from 128 to 512. Larger embeddings increase the memory required for hidden representations and intermediate activations. Across the tested dimensions, AA-MP requires less allocated GPU memory than both standard BF16 AMP and the available FP32 reference, with the absolute memory advantage increasing at larger embedding dimensions. The hot-step timing results are less uniform across reduced-precision configurations and do not establish a consistent timing advantage of AA-MP over AMP at every embedding dimension. Because the FP32 reference and reduced-precision results originate from different scaling workflows, these measurements are interpreted as resource-scaling trends rather than strictly matched speedup estimates.

\FloatBarrier

\Needspace{8\baselineskip}

\section{Conclusion}
\label{sec:conclusion}

This work proposed architecture-aware mixed precision (AA-MP) for replica-exchange training of geometry-aware neural operators. By retaining FP32 computation in the cold-replica and employing controlled BF16 computation with selected FP32 states and reductions in the hot-replica, AA-MP assigns numerical precision according to replica roles and network architecture. Experiments on AirfRANS and jet-engine bracket benchmarks demonstrate that AA-MP reduces hot-phase GPU memory and step time relative to FP32 and standard AMP while maintaining comparable predictive accuracy. Scaling studies show a growing absolute GPU-memory advantage over AMP at larger tested workloads. Precision-allocation ablation and Naive-BF16 baseline further demonstrate that uncontrolled precision reduction degrades predictive accuracy, supporting the need for controlled mixed-precision computation. A separate conformal-UQ evaluation achieves mean spatial coverage near the nominal target with narrower mean prediction intervals, although case-level coverage varies. These findings indicate that explicit precision control can improve computational efficiency beyond standard AMP for the evaluated geometry-aware neural operators.

Future work will extend AA-MP implementation to physics-informed ArGEnT, where higher order derivatives required for PDE residual evaluation increase substantial computational costs, to investigate whether controlled mixed precision can improve computational efficiency while preserving predictive accuracy and physical consistency. Within this controlled mixed precision framework, We will further integrate uncertainty quantification with active learning to guide the selection of informative training samples, particularly for out-of-distribution geometries, and evaluate their effects on uncertainty reliability and data efficiency.

\section*{Acknowledgement}

This work was supported by Sandia National Laboratories under the Interlaboratory (IL) Laboratory Directed Research and Development (LDRD) Program. Sandia National Laboratories is a multi-mission laboratory managed and operated by National Technology \& Engineering Solutions of Sandia, LLC (NTESS), a wholly owned subsidiary of Honeywell International Inc., for the U.S. Department of Energy’s National Nuclear Security Administration (DOE/NNSA) under contract DE-NA0003525. This written work is partially authored by an employee of NTESS. The employee, not NTESS, owns the right, title and interest in and to the written work and is responsible for its contents. Any subjective views or opinions that might be expressed in the written work do not necessarily represent the views of the U.S. Government. The publisher acknowledges that the U.S. Government retains a non-exclusive, paid-up, irrevocable, world-wide license to publish or reproduce the published form of this written work or allow others to do so, for U.S. Government purposes. The DOE will provide public access to results of federally sponsored research in accordance with the DOE Public Access Plan. Guang Lin also acknowledges support from the National Science Foundation (DMS-2533878, DMS-2053746, DMS-2134209, ECCS-2328241, CBET-2347401 and OAC-2311848), and U.S.~Department of Energy (DOE) Office of Science Advanced Scientific Computing Research program under the "Uncertainty Quantification for Multifidelity Operator Learning (MOLUcQ)" project (Project No. 81739), DE-SC0023161, the SciDAC LEADS Institute, and DOE–Fusion Energy Science, under grant number: DE-SC0024583.




\clearpage
\appendix
\numberwithin{figure}{section}
\renewcommand{\thefigure}{\Alph{section}.\arabic{figure}}
\section{Additional AirfRANS Conformal UQ Visualizations}
\label{app:uq_visualization}

The following figures complement the FP32/AA-MP result in Fig.~\ref{fig:uq_aamp} by displaying the corresponding spatial uncertainty fields for the other precision configurations on the same held-out AirfRANS case~804. All figures use the same field arrangement: truth, predictive mean, calibrated 95\% interval half-width, and interval-coverage diagnostic. Truth and prediction use common field-specific color limits across methods, and the half-width maps use the same field-specific scales to support direct visual comparison. The coverage categories indicate where the truth falls below, inside, or above the prediction-centered conformal interval. Aggregate 98-case results are reported in Table~\ref{tab:uq_conformal_comparison}.

\begin{figure}[H]
    \centering
    \includegraphics[width=0.95\textwidth]{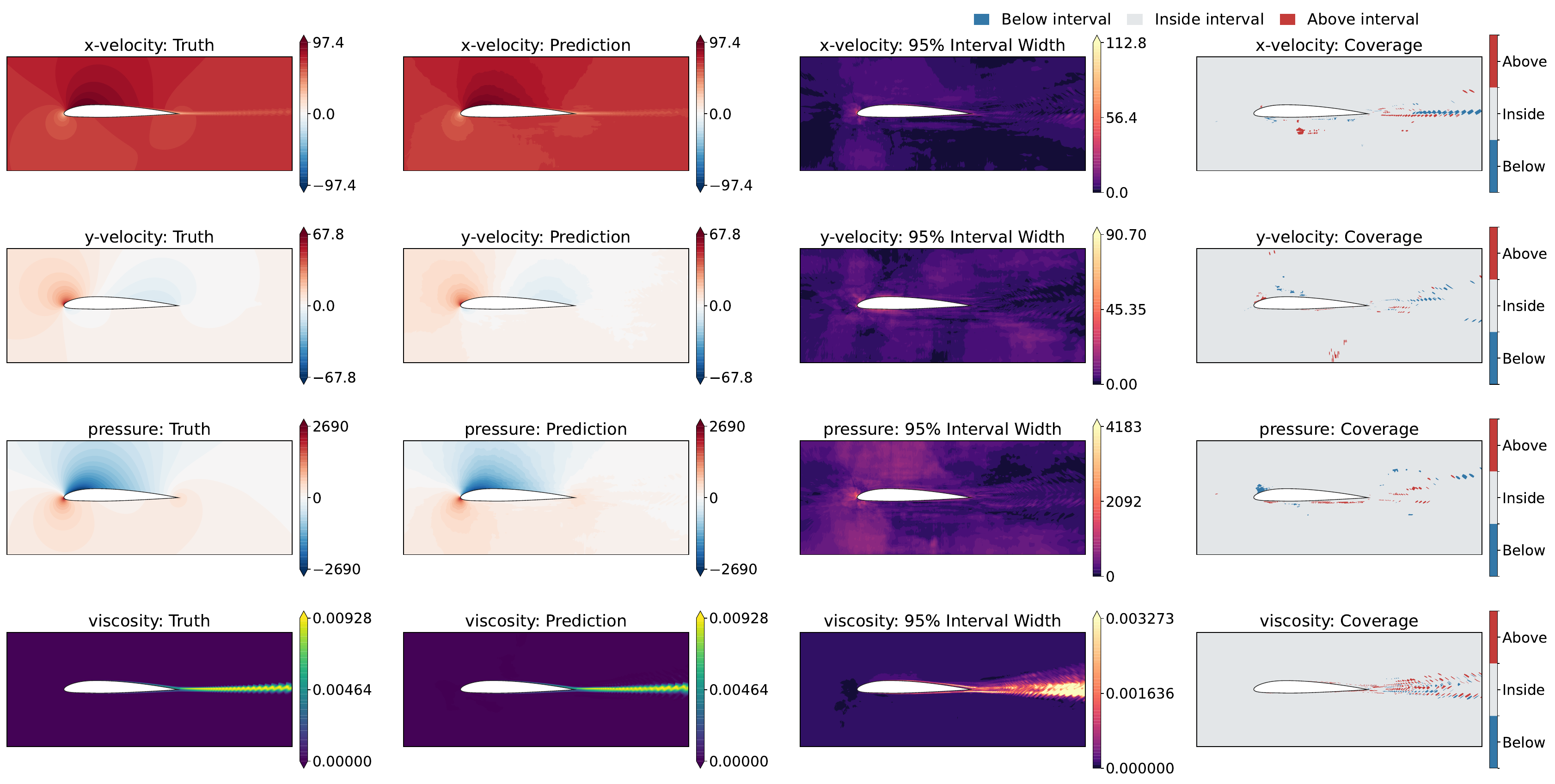}
    \caption{FP32/FP32: spatially resolved post-training conformal UQ for AirfRANS evaluation case~804. For $u$, $v$, $p$, and $\nu_t$ (rows), the columns display truth, predictive mean, calibrated 95\% half-width, and interval coverage. Blue denotes truth below the interval, gray denotes truth inside, and red denotes truth above.}
    \label{fig:uq_fp32}
\end{figure}

\begin{figure}[H]
    \centering
    \includegraphics[width=0.95\textwidth]{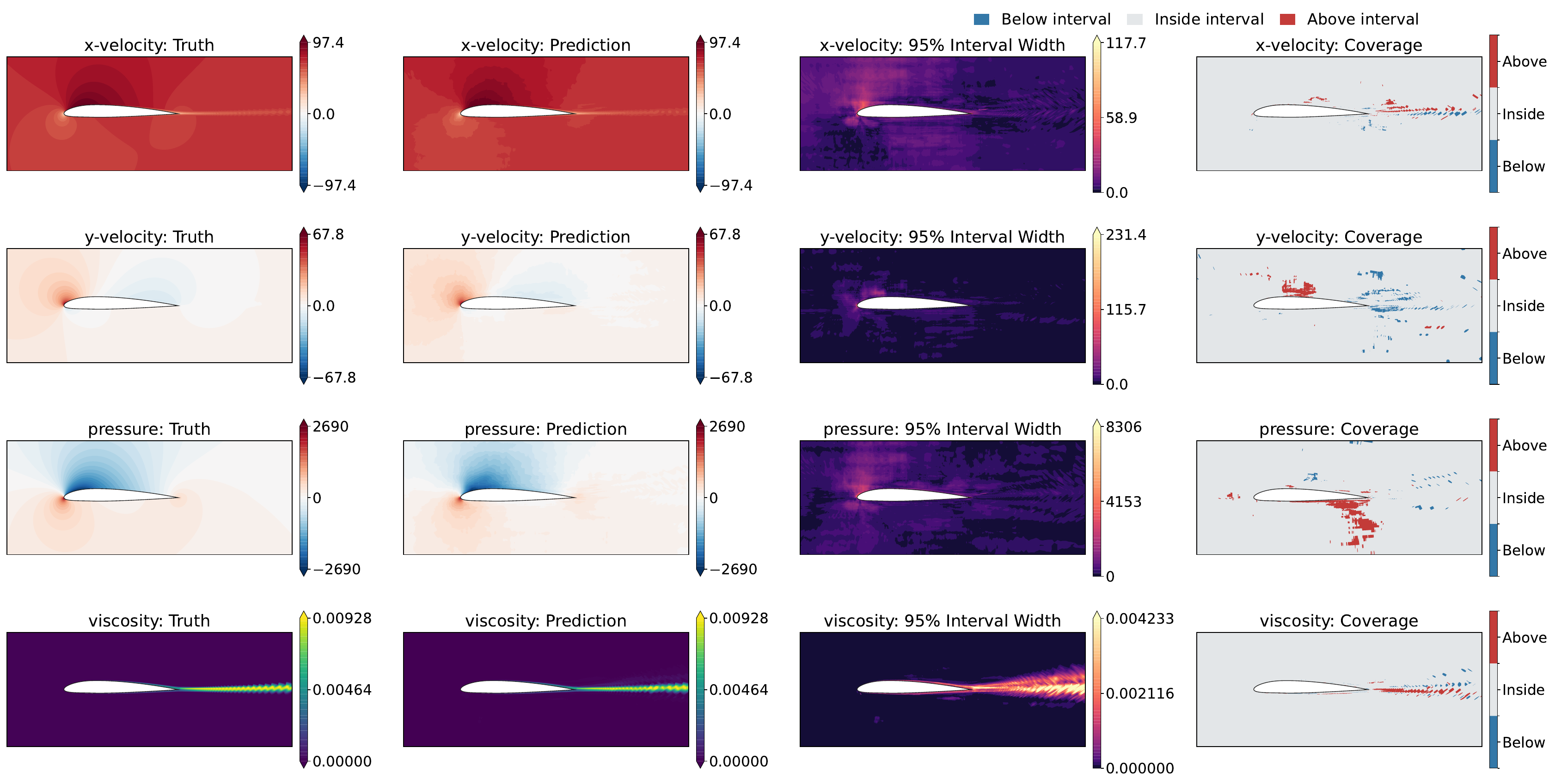}
    \caption{FP32/AMP: spatially resolved post-training conformal UQ for the same AirfRANS evaluation case~804. Panel definitions and categorical coverage colors are identical to those of Fig.~\ref{fig:uq_fp32}.}
    \label{fig:uq_amp}
\end{figure}

\begin{figure}[H]
    \centering
    \includegraphics[width=0.95\textwidth]{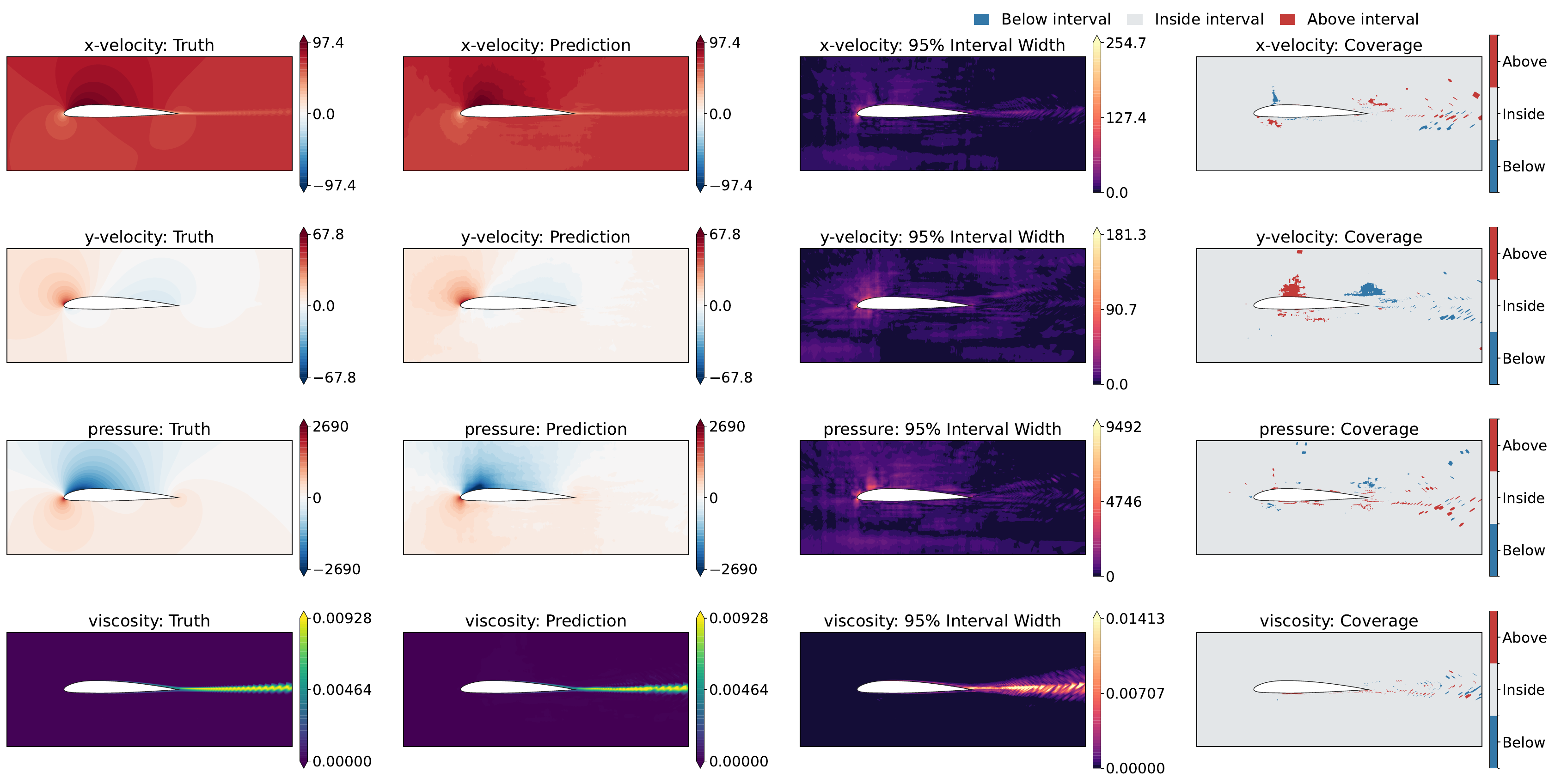}
    \caption{FP32/Naive-BF16: spatially resolved post-training conformal UQ for AirfRANS evaluation case~804. The panels show truth, predictive mean, calibrated 95\% half-width, and categorical interval coverage. Aggregate calibration and interval-score results are reported in Table~\ref{tab:uq_conformal_comparison}.}
    \label{fig:uq_naive_bf16}
\end{figure}
\FloatBarrier

\clearpage
\bibliographystyle{elsarticle-num}
\bibliography{refs}

\end{document}